 \documentclass[tablecaption=bottom,wcp]{jmlr} % W&CP article

\usepackage{booktabs}
\usepackage{parskip}
\usepackage{multirow}
\theorembodyfont{\upshape}
\theoremheaderfont{\scshape}
\theorempostheader{:}
\theoremsep{\newline}

\jmlrproceedings{AABI 2025}{Workshop at the 7th Symposium on Advances in Approximate Bayesian Inference (non-archival), 2025}

\title[Exploring Pseudo-Token Approaches in Transformer Neural Processes]{Exploring Pseudo-Token Approaches in Transformer Neural Processes}

\author{%
  \Name{Jose Lara-Rangel}\thanks{Equal contribution.}\Email{jml224@cam.ac.uk}\\
  \Name{Nanze Chen}\footnotemark[1]\Email{nc630@cam.ac.uk}\\
  \Name{Fengzhe Zhang}\footnotemark[1]\Email{fz287@cam.ac.uk}\\
  \addr Department of Engineering\\
       University of Cambridge
}
\begin{document}

\maketitle

\begin{abstract}
Neural Processes (NPs) have gained attention in meta-learning for their ability to quantify uncertainty, together with their rapid prediction and adaptability. However, traditional NPs are prone to underfitting. Transformer Neural Processes (TNPs) significantly outperform existing NPs, yet their applicability in real-world scenarios is hindered by their quadratic computational complexity relative to both context and target data points. To address this, pseudo-token-based TNPs (PT-TNPs) have emerged as a novel NPs subset that condense context data into latent vectors or pseudo-tokens, reducing computational demands. We introduce the Induced Set Attentive Neural Processes (ISANPs), employing Induced Set Attention and an innovative query phase to improve querying efficiency. Our evaluations show that ISANPs perform competitively with TNPs and often surpass state-of-the-art models in 1D regression, image completion, contextual bandits, and Bayesian optimization. Crucially, ISANPs offer a tunable balance between performance and computational complexity, which scale well to larger datasets where TNPs face limitations.
\end{abstract}

% Keywords may be removed
%\begin{keywords}
%List of keywords
%\end{keywords}

\section{Introduction}
\label{sec:intro}

Function approximation underpins many machine learning problems. Deep Neural Networks (DNNs) are a predominant technique for this, offering efficient predictions after extensive training but struggling to incorporate new data post-training to improve performance further. As alternative, Gaussian Processes (GPs) \citep{rasmussen2006gaussian} provide flexible function regression by conditioning on observed data to infer the underlying function, facilitating the representation of uncertainty in predictions. However, their cubic computational complexity respect to dataset size limits their scalability for large problems.

In response, Neural Processes (NPs) \citep{garnelo2018conditional, garnelo2018neural} aim to combine the advantages of DNNs and GPs, offering uncertainty quantification, fast prediction, and continual learning. However, early NP models often struggle with poor context data fitting and suboptimal performance. Transformer Neural Processes (TNPs) \citep{nguyen2022transformer, vaswani2017attention} improve NP performance across tasks but face scalability issues due to their quadratic complexity in both training and querying. To mitigate this, pseudo-token-based TNPs (PT-TNPs) use latent vectors to efficiently summarize context data, reducing computational and memory costs. A key PT-TNP variant, Latent Bottlenecked Attentive Neural Processes (LBANPs) \citep{feng2022latent}, further reduces complexity, achieving linear query time relative to target points and independence from context size.

Based on \cite{lee2019set}, we implement the Induced Set Attentive Neural Process (ISANP) and introduce ISANP-2 with an innovative query phase. Both ISANPs encode the context dataset into a fixed set of latent vectors, which are accessed via cross-attention with target data for predictions. Through extensive empirical analysis on 1D regression, image completion, contextual bandits, and Bayesian optimization, ISANPs outperformed LBANPs and previous NPs variants, showing a comparable performance to TNPs. Moreover, ISANPs also offer a tunable balance between performance and computational efficiency based on the number of latent vectors and exhibit capacity to scale to large datasets, addressing the limitations faced by traditional attention-based NP variants.

\section{Background}
% \subsection{Meta-Learning with Uncertainty Quantification}
% In meta-learning, we assume an unknown underlying distribution of functions, denoted as $F$. During training, a fixed number of functions $f_j\sim F$ are sampled, each paired with a finite context dataset ${(x_i, y_i)}{i=1}^N$ and a distinct target dataset ${(x_i, y_i)}{i=1}^M$, where $N$ and $M$ are the sizes of the context and target sets, respectively. Initially, the context dataset is used to perform an update to the model with the type of update varying across different models. Then, model's performance is evaluated on the target dataset and the update rule adjusted based on this evaluation. The testing phase also includes testing the model on unseen functions drawn from $F$, assessing its generalization capability through predictions on a new target dataset $\mathcal{D}_{\text{target}}$ given a small context dataset $\mathcal{D}_{\text{context}}$. 

% For this work, we focus on meta-learning models that can make predictions with uncertainty quantification. Specifically, these models predict a joint distribution over the target dataset, enabling uncertainty estimation for each prediction. For example, assuming a Gaussian predictive distribution, the model outputs a mean $\mu_j$ and standard deviation $\sigma_j$ for each input $x_j$, where $\sigma_j$ represents the prediction's uncertainty.

\subsection{Neural Processes}

% Neural Process (NP) models are typically categorized into two distinct types: predictive NPs and generative NPs. Generative NPs adhere to the Kolmogorov Extension Theorem \citep{oksendal2013stochastic}, satisfying both exchangeability and consistency criteria, thus constituting properly defined Stochastic Processes. Conversely, predictive NPs, which make predictions based on a context dataset of labeled observations, fulfill the exchangeability criterion but not consistency. Predictive NPs have shown effectiveness in applications like 1D regression, Image Completion, and few-shot image classification. However, the absence of consistency in predictive NPs may lead to challenges in dynamic contexts requiring sequential decision updates, potentially causing discrepancies in uncertainty estimation as new data is incorporated \citep{xu2024deep}. Our work, however, will concentrate on predictive NPs, aligning with our focus on prior benchmarks including CNPs, CANPs, TNPs, LBANPs, among others.
\begin{table}[t]
\centering
\caption{Computational complexity of models with respect to the number of context datapoints $(N)$, number of target datapoints $(M)$ and number of latent vectors $(L)$.}
\label{tab: computational complexity}
\begin{tabular}{lccc}
\toprule
Method & Training Step & Condition & Query\\
\midrule
CNP \citep{garnelo2018conditional} & $\mathcal{O}(N+M)$ & $\mathcal{O}(N)$ & $\mathcal{O}(M)$ \\
CANP \citep{kim2019attentive} & $\mathcal{O}(N^2+NM)$ & $\mathcal{O}(N^2)$ & $\mathcal{O}(NM)$\\
NP \citep{garnelo2018neural} & $\mathcal{O}(N+M)$ & $\mathcal{O}(N)$ & $\mathcal{O}(M)$ \\
ANP \citep{kim2019attentive} & $\mathcal{O}(N^2+NM)$ & $\mathcal{O}(N^2)$ & $\mathcal{O}(NM)$ \\
TNP-D \citep{nguyen2022transformer} & $\mathcal{O}((N+M)^2)$ & {---} & $\mathcal{O}((N+M)^2)$ \\
\midrule
LBANP \citep{feng2022latent} & $\mathcal{O}((N +M +L)L)$ & $\mathcal{O}((N+L)L)$ & $\mathcal{O}(ML)$ \\
\textbf{ISANP} & $\mathcal{O}((2N +M)L)$  & $\mathcal{O}(2NL)$ & $\mathcal{O}(ML)$ \\
\textbf{ISANP-2 (ours)} & $\mathcal{O}((2L +M)N)$ & $\mathcal{O}(2NL)$ & $\mathcal{O}(MN)$\\
\bottomrule
\end{tabular}
\vspace{-0.6em}
\end{table}

NPs are meta-learning models that provide predictions along with uncertainty quantification. We focus on NP variants that generate predictions directly conditioned on a context dataset. Given a labeled context dataset $\{(x_i,y_i)\}_{i=1}^N$ and an unlabeled target dataset $\{x_j\}_{j=1}^M$, the NP framework comprises three components. The \emph{encoder} transforms each pair $(x_i,y_i)$ into a representation $r_i = h_\theta(x_i,y_i)$, where $h_\theta$ is a neural network with parameters $\theta$. The \emph{aggregator} combines these representations into a summary $r_C = a(\{r_i\}_{i=1}^N)$; a common choice is the mean $r_C = \frac{1}{N}\sum_{i=1}^N r_i$, ensuring order invariance (alternatives include self-attention layers \citep{kim2019attentive}). The \emph{conditional decoder} $g_\phi$ then integrates $r_C$ with each target input $x_j$ to produce predictions. 
% Assuming prediction independence, the overall predictive distribution is 
% \[
% p(y_{1:M}\mid x_{1:M},\mathcal{D}_{\text{context}})=\prod_{j=1}^M p(y_j\mid x_j,r_C).
% \]

NPs provide more efficient predictions than GPs (see Table~\ref{tab: computational complexity} for details). Although NPs require only a single conditional step for predictions, each query is processed independently, making the optimization of the query step’s complexity crucial for enhancing NP efficiency.

\subsection{Transformer Neural Processes}
TNPs \citep{nguyen2022transformer} leverage transformer-based architectures to process both context and target data points. Instead of assuming independent predictions, TNPs model the joint distribution of all target points. The predictive distribution is represented as
\begin{equation}
p(y_{1:M}|x_{1:M}, \mathcal{D}_{\text{context}}) := p(y_{1:M}|r_{C, T})
\end{equation}
where $r_{C, T}$ is the aggregated representation from both context and target data points through multiple self-attention layers, using a masking mechanism ensuring context and target datapoints only attend to context datapoints. The embeddings of the target data are passed to the predictor to make predictions. Embeddings are computed simultaneously for both context and target datasets, removing the need for a separate conditioning step. However, this requires recomputing embeddings for each prediction, leading to a computational complexity of $\mathcal{O}((N+M)^2)$ that limits TNPs' applicability on large-scale datasets.

% TNPs have three variants: Autoregressive TNP (TNP-A), Diagonal TNP (TNP-D), and Non-Diagonal TNP (TNP-ND). TNP-A makes autoregressive predictions at target points, conditioning on the context dataset, but is computationally intensive. TNP-D assumes all predictions are independent, while TNP-ND balances TNP-D's efficiency with TNP-A's expressivity, modeling the predictive distribution as a multivariate normal with a non-diagonal covariance matrix.

\section{Pseudo-tokens Based Transformer Neural Processes}
PT-TNPs are a novel subclass within the NPs family that leverage Transformer-like architectures and efficient attention techniques to simplify computational complexity by condensing context dataset information into a minimal set of latent vectors. This family includes Induced Points Neural Processes (ISNPs) \citep{rastogi2022semi} and LBANPs \citep{feng2022latent}, alongside our contribution, ISANPs. ISNPs extend the Non-parametric Transformer (NPT) \citep{kossen2021self} and, similar to our work, utilize the Set Transformer architecture \citep{lee2019set} to reduce computational demands. However, different to ISANPs, ISNPs employ two sets of latent vectors and a unique cross-attention methodology to summarize context dataset information for predictions.

\section{Methodology}

\subsection{Induced Set Attentive Neural Process}
\begin{figure}[t]
    \centering
    \includegraphics[width=0.9\linewidth]{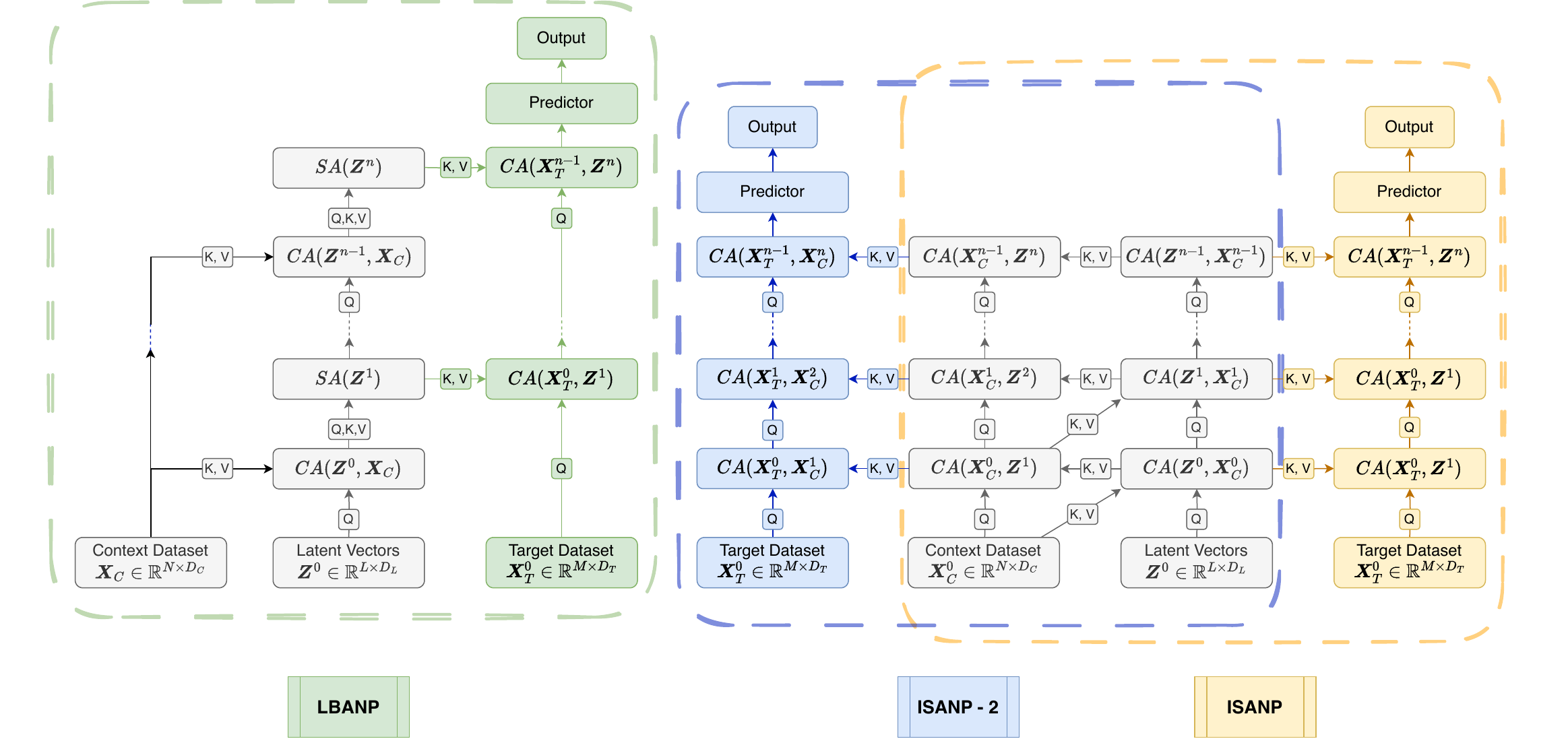}
    \caption{LBANP architecture together with the two proposed ISANPs architectures. $CA$ stands for cross-attention and $SA$ stands for self-attention.}
    \label{fig: model structures}
\vspace{-0.5em}
\end{figure}

Building on insights from LBANPs \citep{feng2022latent} and Set Transformer \citep{lee2019set}, we introduce ISANPs, a novel NP with a transformer-based architecture that reduces query complexity and outperforms attention-based NPs on various tasks. We present two variants: ISANP and ISANP-2. Both use a set of latent vectors, sized $L\times D^L$, to encode context dataset information. The hyperparameter $L$ allows to manage the information bottleneck size, enabling a balance between performance and computational complexity.

\textbf{Conditioning Phase} Both ISANP variants have the same conditioning phases. Instead of combining cross-attention with subsequent self-attention over latent vectors, this phase incorporates an additional cross-attention between the context dataset and latent vectors. Latent Embeddings (LEMB) are initialized as learnable parameters. The conditioning phase for both variants is formalized as follows, given $\text{CEMB}_0 = \mathcal{D}_{\text{context}}$:
\begin{equation*}
\text{LEMB}_i = \text{CA}(\text{LEMB}_{i-1}, \text{CEMB}_{i-1}), \quad\text{CEMB}_i = \text{CA}(\text{CEMB}_{i-1}, \text{LEMB}_{i})
\end{equation*}
% \begin{align*}
% \text{CEMB}_0 &= \mathcal{D}_{\text{context}}\\
% \text{LEMB}_i &= \text{CrossAttention}(\text{LEMB}_{i-1}, \text{CEMB}_{i-1})\\
% \text{CEMB}_i &= \text{CrossAttention}(\text{CEMB}_{i-1}, \text{LEMB}_{i})
% \end{align*}
Utilizing two cross-attentions, the conditioning phase's total computational complexity reaches $\mathcal{O}(2NL)$. As conditioning occurs once per context dataset, this increase does not substantially impact model efficiency.

\textbf{Query Phase} ISANP employs multiple cross-attention stages to extract context dataset information from latent vectors resulting in a computational complexity matches of $\mathcal{O}(ML)$. ISANP-2 introduces a different query phase, akin to \citep{feng2022latent}, where cross-attention between the target dataset and context embeddings is performed, with $\text{QEMB}_0 = \text{QUERY}$:
\begin{equation*}
\text{QEMB}_i = \text{CrossAttention}(\text{QEMB}_{i-1}, \text{CEMB}_i),\quad \text{Output} = \text{Predictor}(\text{QEMB}_K)
\end{equation*}
% \begin{align*}
% \text{QEMB}_0 &= \text{QUERY}\\
% \text{QEMB}_i &= \text{CrossAttention}(\text{QEMB}_{i-1}, \text{CEMB}_i)\\
% \text{Output} &= \text{Predictor}(\text{QEMB}_K)
% \end{align*}
Here, the computational complexity during the query of ISANP-2 is $\mathcal{O}(NM)$, higher than the computational complexity of ISANP but still lower than TNPs.

% To summarize, the conditioning phases of ISANPs and LBANPs differ slightly due to the use of two cross-attentions between CEMB and LEMB. This difference leads to a higher computational complexity for ISANPs compared to LBANPs. During the query phase, however, ISANP and LBANP exhibit the same computational complexities. In the case of ISANP-2, the computational complexity increases as cross-attention is applied between the context and target datasets. A key advantage of both LBANP and ISANP models lies in their lower memory requirements. This efficiency stems from storing all context dataset information within a fixed-size set of latent vectors, ensuring a constant memory requirement. Such an approach is particularly advantageous for data privacy purposes, as it allows to discarding of the context dataset.

\section{Experiments}
\begin{table}[t]
\caption{Log-Likelihood for 1D meta-regression experiments using 5 seeds (higher is better).}
\label{1_d_table_partial}
\centering
\begin{tabular}{lccc}
\toprule
Method  & RBF & Matérn 5/2 & Periodic \\
\midrule
TNP-D & 1.39 $\pm$ 0.00 & 0.95 $\pm$ 0.01 & -3.53 $\pm$ 0.37\\
%EQTNP & 1.32 $\pm$ 0.01 & 0.92 $\pm$ 0.01 & {---}\\
%LBANP (8) & 1.20 $\pm$ 0.02 & 0.75 $\pm$ 0.02 & - \\
%LBANP (128) & 1.27 $\pm$ 0.02 & 0.85 $\pm$ 0.02 & -\\
\midrule
LBANP (8) & 1.19 $\pm$ 0.02 & 0.76 $\pm$ 0.02 & -2.67 $\pm$ 0.01\\
LBANP (128) &  1.24 $\pm$ 0.02 & 0.88 $\pm$ 0.02 & -1.93 $\pm$ 0.01\\
%\hline
ISANP (8) & 1.26 $\pm$ 0.02 & 0.81 $\pm$ 0.02 & -2.85 $\pm$ 0.01\\
ISANP (128) & 1.34 $\pm$ 0.02 & 0.90 $\pm$ 0.02 & -3.69 $\pm$ 0.01 \\
ISANP-2 (8) & 1.20 $\pm$ 0.01 & 0.75 $\pm$ 0.01 & -6.53 $\pm$ 0.07\\
ISANP-2 (128) & 1.19 $\pm$ 0.01 & 0.82 $\pm$ 0.01 & -9.99 $\pm$ 0.07\\
\bottomrule
\end{tabular}
\vspace{-0.34em}
\end{table}
We evaluate and compare ISANPs against LBANP, traditional NP variants and TNPs in four different tasks commonly used to benchmark NP models \citep{garnelo2018conditional, nguyen2022transformer,kim2019attentive}. Here, we present the results for 1D meta-regression and image completion, and include results for contextual bandits and Bayesian optimization in Appendix \ref{apd:first}. For TNPs, we focus on TNP-D as it preserves the independence assumption, modeling the probabilistic predictive distribution identical to traditional NPs, offering a fair comparison. We set $L=8$ and $L=128$ ISANPs, and analyze how the number of latent vectors impacts the models performance and scalability to large context or target datasets in terms of memory and time complexity. We present ablation studies in Appendix \ref{apd:ablation_studies}.

\vspace{-0.4em}
\subsection{1D Meta-Regression}

The model is given a set of $N$ context datapoints and has to make predictions on a set of $M$ target datapoints, both sets drawn from the same underlying unknown function $f$. In this case, the function is sampled from a GP prior, $f_i \sim \text{GP}(m, k)$, using $m(x) = 0$ and an RBF kernel. In each training epoch, a batch of 16 functions is used, with hyperparameters sampled uniformly at random for each task as follows: $l \sim \mathcal{U}[0.6, 1.0)$, $\sigma_f \sim \mathcal{U}[0.1, 1.0)$, $N \sim \mathcal{U}[3, 47)$, and $M \sim \mathcal{U}[3, 50 - N)$. The models are evaluated during test time according to the log-likelihood of target datapoints sampled from GPs with RBF, Matérn 5/2, and Periodic kernels. The sizes of the context and target sets are sampled similarly as in training.

\begin{table}[t]
\caption{Log-Likelihood for image completion experiments (higher is better). Dashes indicate methods that could not be run due to memory or computational constraints.}
\label{image_comp_table_partial}
\centering
\begin{tabular}{lcccc}
\toprule
\multirow{2}{*}{Method} & \multicolumn{2}{c}{CelebA} & \multicolumn{2}{c}{EMNIST}\\
& 32x32 & 64x64 & Seen (0-9) & Unseen (10-46)\\ 
\midrule
TNP-D & 3.89 $\pm$ 0.01 & 5.41 $\pm$ 0.01 & 1.46 $\pm$ 0.01 & 1.31 $\pm$ 0.00\\
\midrule
LBANP (8) & 3.50 & 4.44 & 1.33 & 1.04 \\
LBANP (128) & 3.86 & - & 1.39 & 1.17 \\
ISANP (8) & 3.58 & 4.74 & 1.39 & 1.10 \\
ISANP (128) & 3.86 & - & 1.42 & 1.17 \\
ISANP-2 (8) & 3.82 & 5.24 & 1.43 & 1.21 \\
ISANP-2 (128) & 3.78 & - & 1.43 & 1.22 \\
\bottomrule
\end{tabular}
\vspace{-0.6em}
\end{table}

\textbf{Result:}
Table \ref{1_d_table_partial} shows that both ISANPs surpass LBANP and achieve competitive results with TNP-D with only 8 latent vectors. Increasing the number of latent vectors to 128 further reduces the performance gap. Notably, both ISANPs with 8 latent vectors achieved competitive results with LBANP with 128 latent vectors. In this task, ISANP outperformed ISANP-2. For functions sampled by different NP models and complete experiment results, which shows that both ISANP models outperform previous NP models, see Appendix \ref{apd:1d_meta_reg}.

\vspace{-0.6em}
\subsection{Image Completion}
A subset of pixel values of an image are passed to the model to complete it by predicting the remaining pixels, which is equivalent to a 2D meta-regression \citep{garnelo2018neural}. We consider the EMNIST \citep{cohen2017emnist} and CelebA \citep{liu2015deep} datasets. CelebA contains coloured images of celebrity faces down-sampled to $32\times 32$ and $64\times 64$. Values are re-scaled so that $x \in [-1,1]$ and $y \in [-0.5,0.5]$. The subsets of pixels for context and target datapoints are selected at random with $N \sim \mathcal{U}[3, 797)$ and $M \sim \mathcal{U}[3, 800-N)$ for CelebA64. For details on EMNIST experiments and complete experiment results, see Appendix \ref{apd:img_completion}.

\begin{figure}[t]
  \centering
  \includegraphics[width=0.9\linewidth]{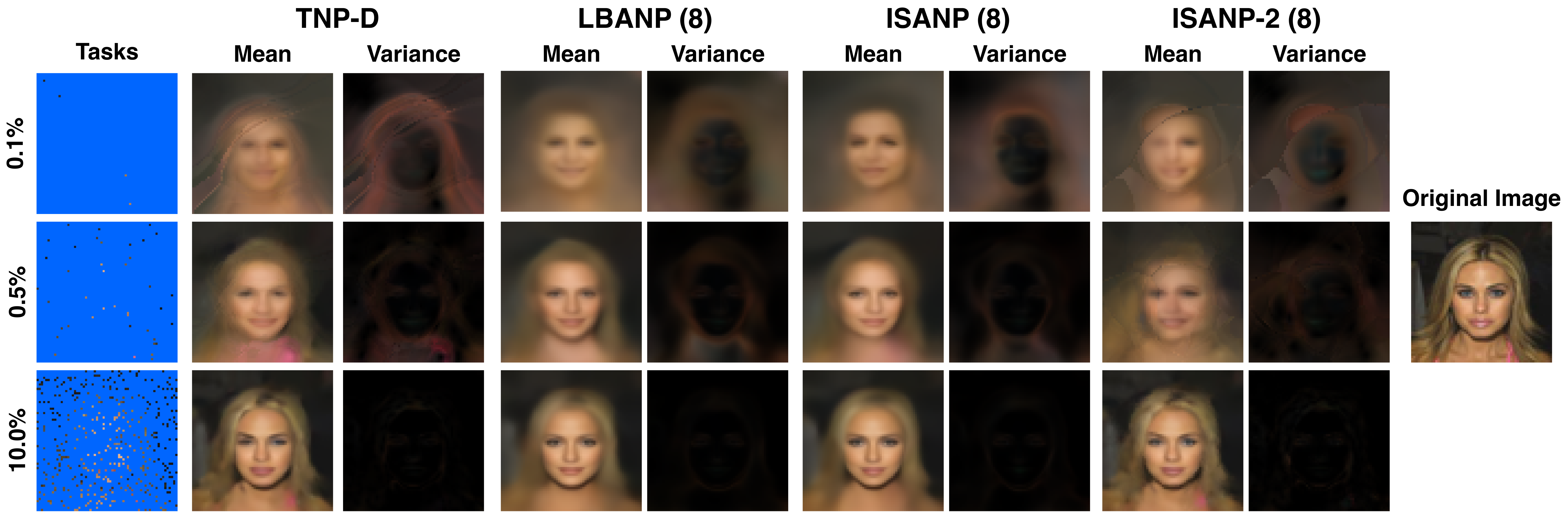}
  \caption{Model performance on CelebA64 using 8 latent vectors.}
\label{fig:celeba_performance}
\vspace{-0.22em}
\end{figure}

\textbf{Result:}
As shown in Table \ref{image_comp_table_partial}, both ISANP models surpass the performance of previous NP models using only 8 latent vectors and approach the performance of TNP-D. Specifically, ISANP-2 achieves performance that is competitive with and comparable to TNP-D. This is further illustrated in Figures \ref{fig:celeba_performance} and \ref{fig:emnist_performance}, where ISANP-2 reconstructed images of remarkable quality with a low proportion of context pixels, and almost indistinguishable from the image resulting from TNP-D, and exhibiting the lowest variance among the subquadratic models presented, followed by ISANP and LBANP. Moreover, increasing the number of latent vectors improves ISANPs performance on both datasets, enabling it to achieve competitive results with TNP-D. However, due to limitations in computational resources, these improvements could not be thoroughly verified. Also, as in previous studies \citep{feng2022latent}, some attention-based NP variants could not be trained due to their high computational demands, underscoring the advantages of ISANPs in terms of scalability and performance.

Notably, while ISANP outperforms ISANP-2 in 1D regression, the opposite is true for image completion. We hypothesize that ISANP-2 benefits from its ability to interact with a large context dataset in tasks where attending to various datapoints that are related across the context is advantageous, such as in image completion. Conversely, in simpler contexts, like 1D regression, the latent vectors alone are adequate for capturing the necessary relationships and higher-order interactions of the context dataset required for accurate predictions.

\section{Conclusions and Future Work}
TNPs, despite its superior performance, face limitations for practical deployment due to their quadratic computational complexity. In response, this work introduces ISANP and ISANP-2, two novel PT-TNPs models, which through extensive evaluations on 1D regression, image completion, contextual bandits and Bayesian optimization show to not only outperform other NP models, including LBANP, but also present a competitive performance with TNPs while allowing to balance performance and computational demand. Additionally, our ablation studies show that ISANPs can scale to larger context dataset sizes, demonstrating the feasibility of improving performance within computational constraints.

Although computational resources constrained our experiments, our results and ablation studies suggest ISANPs' performance would at least match that of LBANPs. Testing with more latent vectors and complex tasks is essential for further comparison with other NPs. Future work could explore ISANPs in higher dimensions, such as Bayesian optimization, or with more latents in image completion, and compare them with TNPs for deeper insights.

\bibliography{references}

\newpage
\appendix

\section{Additional Experimental Results}
\label{apd:first}

\subsection{1D Meta-Regression}
\label{apd:1d_meta_reg}

\begin{figure}[ht]
  \floatconts
  {fig:1d_regression}
  {\caption{1D Meta-Regression sample functions produced by different models.}}
  {%
    \subfigure[CNP]{\label{fig:1d_cnp}%
      \includegraphics[width=0.32\linewidth]{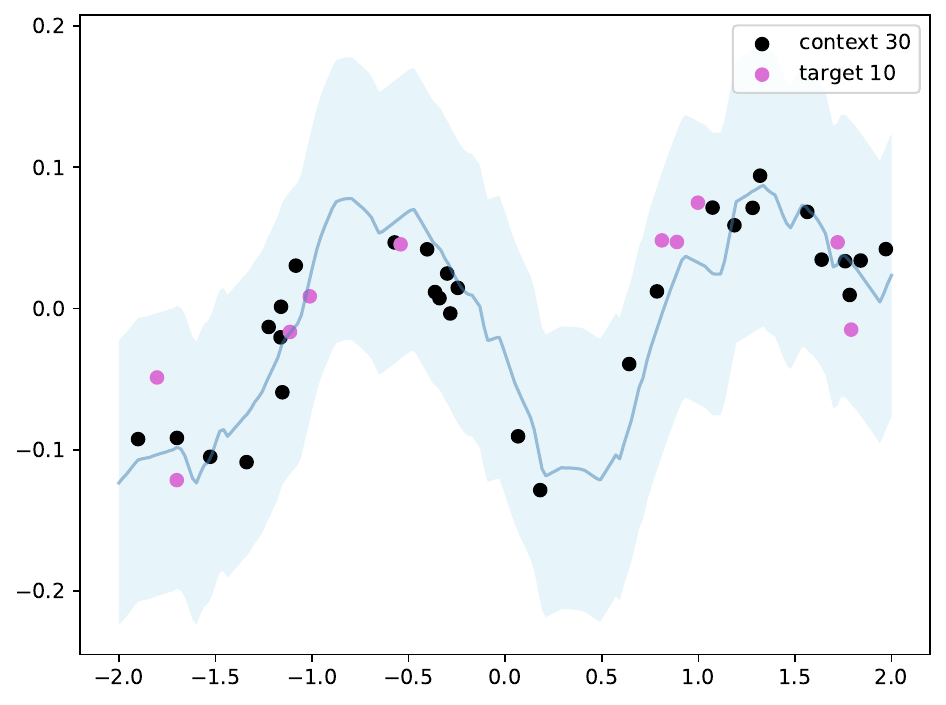}}
    \hfill
    \subfigure[CANP]{\label{fig:1d_canp}%
      \includegraphics[width=0.32\linewidth]{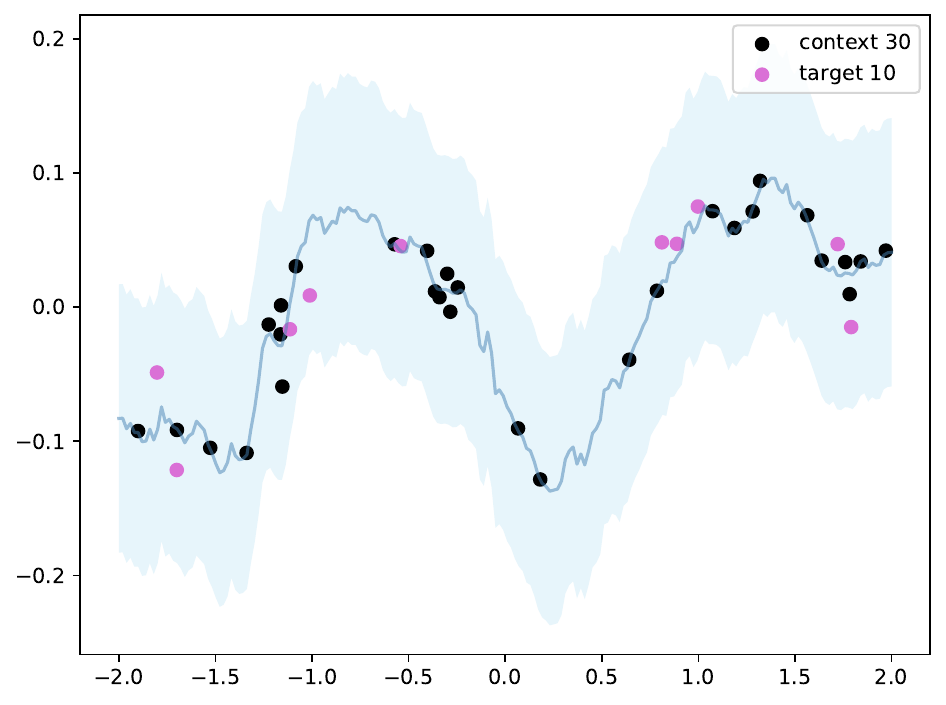}}
    \hfill
    \subfigure[TNP-D]{\label{fig:1d_tnpd}%
      \includegraphics[width=0.32\linewidth]{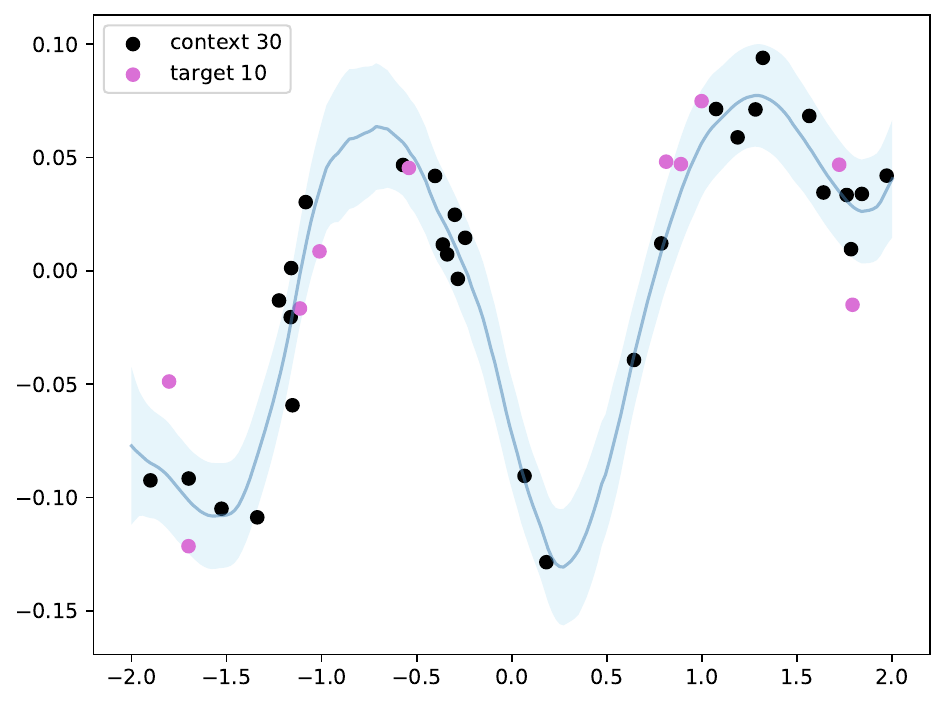}}
    
    \subfigure[LBANP (8)]{\label{fig:1d_lbanp}%
      \includegraphics[width=0.32\linewidth]{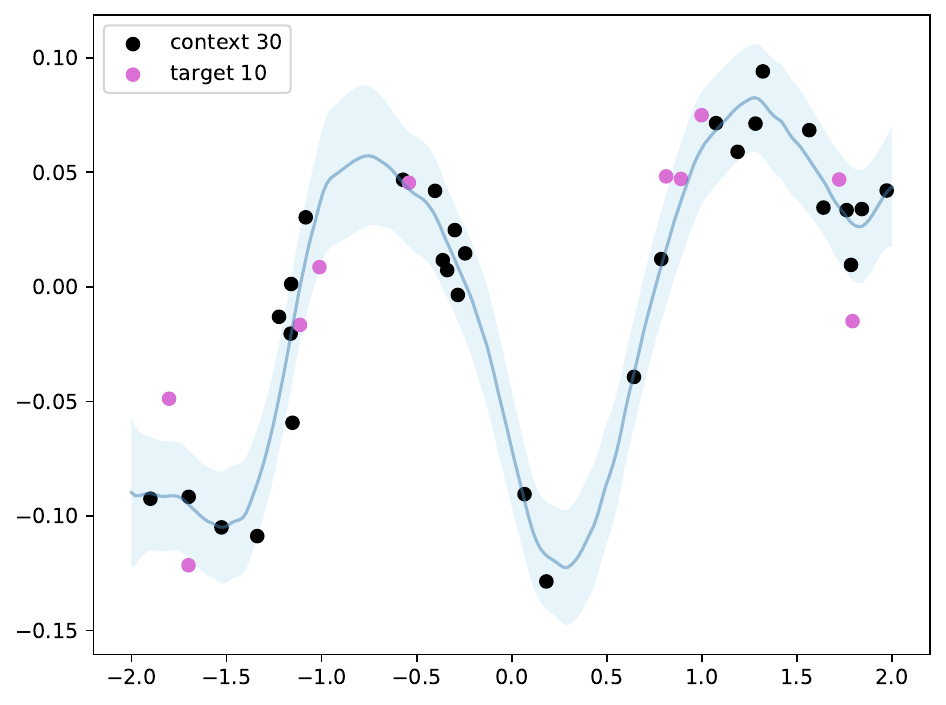}}
    \hfill
    \subfigure[ISANP (8)]{\label{fig:1d_isanp}%
      \includegraphics[width=0.32\linewidth]{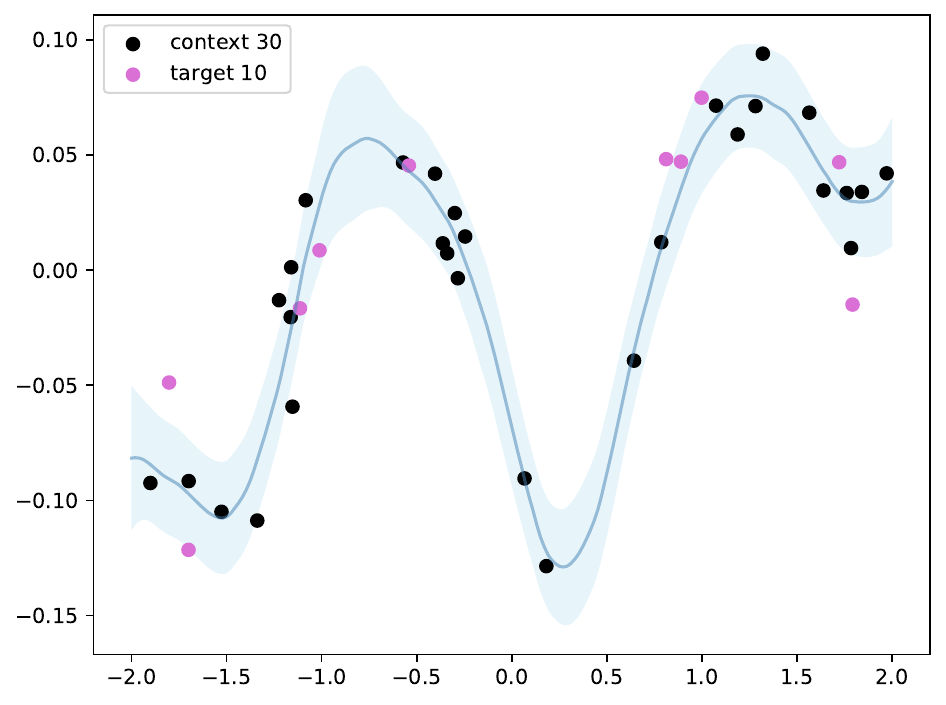}}
    \hfill
    \subfigure[ISANP-2 (8)]{\label{fig:1d_isanp2}%
      \includegraphics[width=0.32\linewidth]{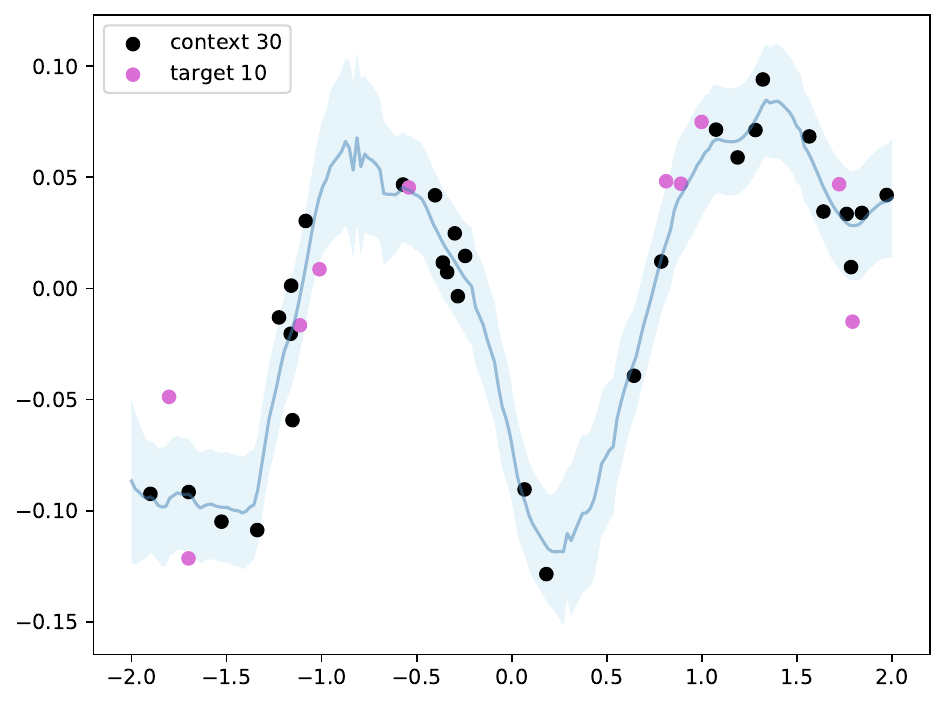}}
  }
\end{figure}

\begin{table}[t]
\caption{Log-Likelihood from complete 1D meta-regression experiments using 5 seeds (higher is better).}
\label{1_d_table}
\centering
\begin{tabular}{lccc}
\toprule
Method  & RBF & Matérn 5/2 & Periodic \\
\midrule
CNP & 0.26 $\pm$ 0.02 & 0.04 $\pm$ 0.02 & -1.40 $\pm$ 0.02 \\
CANP & 0.79 $\pm$ 0.00 & 0.62 $\pm$ 0.00 & -7.61 $\pm$ 0.16\\
NP & 0.27 $\pm$ 0.01 & 0.07 $\pm$ 0.01 & -1.15 $\pm$ 0.04\\
ANP & 0.81 $\pm$ 0.00 & 0.63 $\pm$ 0.00 & -5.02 $\pm$ 0.21\\
BNP & 0.38 $\pm$ 0.02 & 0.18 $\pm$ 0.02 & -0.96 $\pm$ 0.02\\
BANP & 0.82 $\pm$ 0.01 & 0.66 $\pm$ 0.00 & -3.09 $\pm$ 0.14\\
TNP-D & 1.39 $\pm$ 0.00 & 0.95 $\pm$ 0.01 & -3.53 $\pm$ 0.37\\
%EQTNP & 1.32 $\pm$ 0.01 & 0.92 $\pm$ 0.01 & {---}\\
%LBANP (8) & 1.20 $\pm$ 0.02 & 0.75 $\pm$ 0.02 & - \\
%LBANP (128) & 1.27 $\pm$ 0.02 & 0.85 $\pm$ 0.02 & -\\
TNP-ND & 1.46 $\pm$ 0.00 & 1.02 $\pm$ 0.00 & -4.13 $\pm$ 0.33 \\
TNP-A & 1.63 $\pm$ 0.00 & 1.21 $\pm$ 0.00 & -2.26 $\pm$ 0.17\\
\midrule
LBANP (8) & 1.19 $\pm$ 0.02 & 0.76 $\pm$ 0.02 & -2.67 $\pm$ 0.01\\
LBANP (128) &  1.24 $\pm$ 0.02 & 0.88 $\pm$ 0.02 & -1.93 $\pm$ 0.01\\
%\hline
ISANP (8) & 1.26 $\pm$ 0.02 & 0.81 $\pm$ 0.02 & -2.85 $\pm$ 0.01\\
ISANP (128) & 1.34 $\pm$ 0.02 & 0.90 $\pm$ 0.02 & -3.69 $\pm$ 0.01 \\
ISANP-2 (8) & 1.20 $\pm$ 0.01 & 0.75 $\pm$ 0.01 & -6.53 $\pm$ 0.07\\
ISANP-2 (128) & 1.19 $\pm$ 0.01 & 0.82 $\pm$ 0.01 & -9.99 $\pm$ 0.07\\
\bottomrule
\end{tabular}
\end{table}

Figure \ref{fig:1d_regression} shows sample functions produced by different NP models, given 30 context points and predicting 10 target datapoints from the same function sampled using a GP with an RBF kernel. Each sample function is depicted as a solid blue curve surrounded by a blue area, representing the uncertainty of the predictive distribution over $y$. Although all methods produced diverse sample functions, it is evident that the TNP-D, LBANP, ISANP and ISANP-2 significantly reduce the uncertainty around the predicted mean compared with CNP, CANP, indicating better performance for this task. Specifically, the predicted mean functions obtained with ISANP and ISANP-2 show that the former produces a smoother function and thus achieves better generalization ability, aligning with results in Table \ref{1_d_table}.

\subsection{Image Completion}
\label{apd:img_completion}

\begin{table}[t]
\caption{Log-Likelihood from complete image completion experiments (higher is better). Dashes indicate methods that could not be run due to memory or computational constraints.}
\label{image_comp_table}
\centering
\begin{tabular}{lcccc}
\toprule
\multirow{2}{*}{Method} & \multicolumn{2}{c}{CelebA} & \multicolumn{2}{c}{EMNIST}\\
& 32x32 & 64x64 & Seen (0-9) & Unseen (10-46)\\ 
\midrule
CNP & 2.15 $\pm$ 0.01 & 2.43 $\pm$ 0.00 &  0.73 $\pm$ 0.00 & 0.49 $\pm$ 0.01 \\
CANP & 2.66 $\pm$ 0.01 & 3.15 $\pm$ 0.00 & 0.94 $\pm$ 0.01 & 0.82 $\pm$ 0.01\\
NP & 2.48 $\pm$ 0.02 & 2.60 $\pm$ 0.01 &  0.79 $\pm$ 0.01 & 0.59 $\pm$ 0.01\\
ANP & 2.90 $\pm$ 0.00 & - & 0.98 $\pm$ 0.00 & 0.89 $\pm$ 0.00 \\
BNP & 2.76 $\pm$ 0.01 & 2.97 $\pm$ 0.00 & 0.88 $\pm$ 0.01 & 0.73 $\pm$ 0.01\\
BANP & 3.09 $\pm$ 0.00 & - & 1.01 $\pm$ 0.00 & 0.94 $\pm$ 0.00\\
TNP-D & 3.89 $\pm$ 0.01 & 5.41 $\pm$ 0.01 & 1.46 $\pm$ 0.01 & 1.31 $\pm$ 0.00\\
TNP-ND & 5.48 $\pm$ 0.02 & - & 1.50 $\pm$ 0.00 & 1.31 $\pm$ 0.00\\
TNP-A & 5.82 $\pm$ 0.01 & - & 1.54 $\pm$ 0.01 & 1.41 $\pm$ 0.01\\
EQTNP & 3.91 $\pm$ 0.10 & 5.29 $\pm$ 0.02 & 1.44 $\pm$ 0.00 & 1.27 $\pm$ 0.00\\
\midrule
LBANP (8) & 3.50 & 4.44 & 1.33 & 1.04 \\
LBANP (128) & 3.86 & - & 1.39 & 1.17 \\
ISANP (8) & 3.58 & 4.74 & 1.39 & 1.10 \\
ISANP (128) & 3.86 & - & 1.42 & 1.17 \\
ISANP-2 (8) & 3.82 & 5.24 & 1.43 & 1.21 \\
ISANP-2 (128) & 3.78 & - & 1.43 & 1.22 \\
\bottomrule
\end{tabular}
\end{table}

EMNIST \citep{cohen2017emnist} contains black and white images of handwritten letters and digits with a resolution of $28\times 28$; for training, we used 10 classes. Values are re-scaled so that $x \in [-1,1]$ and $y \in [-0.5,0.5]$, and the subsets of pixels for context and target datapoints are selected at random with $N \sim \mathcal{U}[3, 197)$ and $M \sim \mathcal{U}[3, 200-N)$. Figure \ref{fig:emnist_performance} presents the results from applying ISANPs and other NPs variants. Table \ref{image_comp_table} shows the complete experiment results.

Noteworthy, limited computational resources constrained the scope of our experiments. The image completion task was conducted only once, precluding an assessment of uncertainty in log-likelihoods. The ISANPs model was not evaluated on the $128\times 128$ CelebA dataset due to computational restrictions, however, based on our results we hypothesize its performance would at least match that of LBANPs. Additionally, we only run ISANPs with 8 latent vectors on large datasets such as CelebA $64\times 64$. Given more time and computing resources, it is essential to test the models with more latent vectors.

\begin{figure}[htbp]
  \floatconts
  {fig:emnist_performance}
  {\caption{Model performance on EMNIST using 8 latent vectors.}}
  {%
    \includegraphics[width=0.92\linewidth]{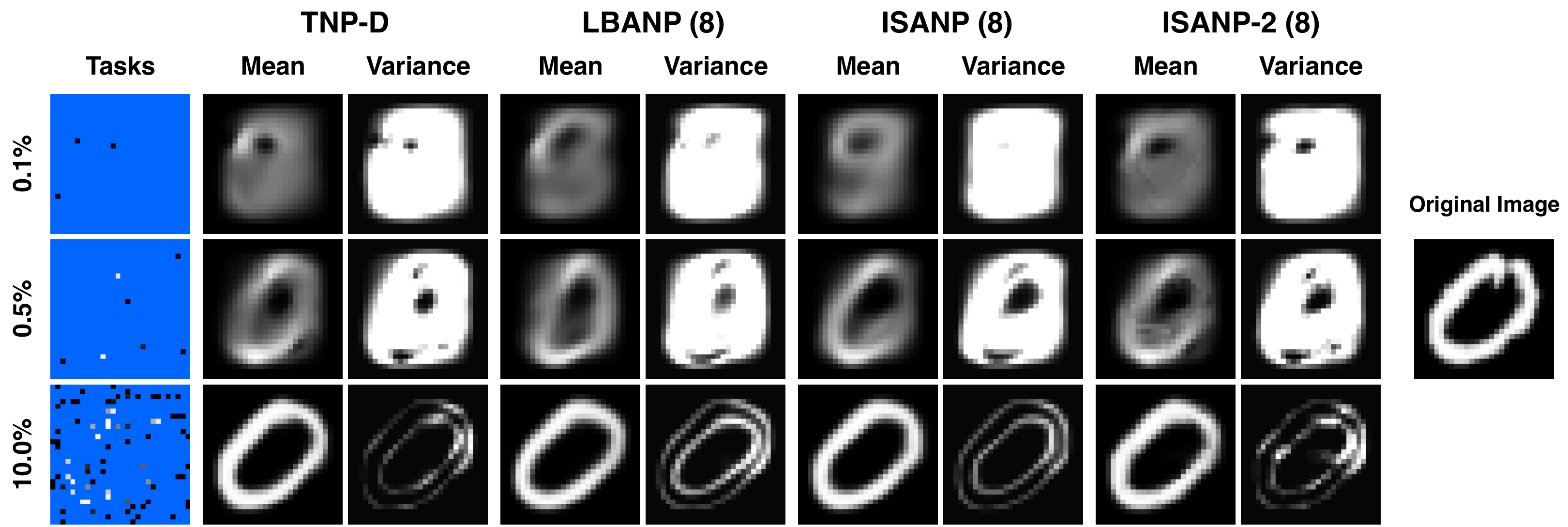}%
  }
\end{figure}

\subsection{Contextual Bandits}
\label{apd:contextual_bandits}

Contextual bandits \citep{riquelme2018deep} is a task often used to test reinforcement learning models' ability to balance exploration and exploitation, which can be used to test NPs' ability to learn from context data points \citep{nguyen2022transformer, feng2022latent, xu2024deep}. In this task, a uniform disk is divided into 5 regions: a central smaller circle as the low-reward region, surrounded by four identical high-reward regions. While the radius of the disk is equal to 1, a scalar $\delta$ controls the radius of the low-reward region. Each region is associated with one action, whose rewards vary in different situations. At each step of this task, a point will be uniformly sampled on the disk. If the sampled point is in the low-reward region, the action associated with the low-reward region will provide reward $r \sim \mathcal{N}(1.2, 0.012)$, and the actions associated with high-reward regions will all provide reward $r \sim \mathcal{N}(1.0, 0.012)$. However, if the sampled point is in one of the high-reward regions, its associated action will provide $r \sim \mathcal{N}(50.0, 0.012)$, while the low-reward regions provide the reward $r \sim \mathcal{N}(1.2, 0.012)$ and the other high-reward regions provide reward $r \sim \mathcal{N}(1.0, 0.012)$. To receive a high reward, the NP models will need to select the correct action based on the sampled point's location with no direct knowledge about the disk or the rules.

In each training iteration, $B = 8$ different $\delta$s are sampled from a uniform distribution $\delta \sim \mathcal{U}(0, 1)$, with each representing a different disk. Then, $M = 50$  target points and $N = 512$ context points will be sampled. Each data point contains $X$ and $R$, where $X$ are the coordinates and $R$ are five actions' reward values. In the circumstances of different $\delta$, $R$ may change for each point. The training objective is to predict the five actions' reward values given the target point coordinates. We then evaluate ISANP, ISANP-2, and baselines on disks with different $\delta$s. For each $\delta$, 2000 evaluation points will be sampled, and 50 different seeds will be used to shuffle them to form 50 groups of sequential data. The performance of models will be evaluated by the average cumulative regret of the 50 experiments.

\begin{table}[t]
\centering
\caption{Contextual Bandit Experiments. Models are evaluated according to cumulative regret (lower is better).}
\label{tab:contextual_bandit_experiment}
\begin{tabular}{lccccc}
\toprule
Method & $\delta$ = 0.7 & $\delta$ =  0.9 & $\delta$ = 0.95 & $\delta$ = 0.99 & $\delta$ = 0.995 \\
\midrule
TNP-D & $1.71 \pm 1.29$ & $2.19 \pm 0.57$ & $2.69 \pm 0.92$ & $3.57 \pm 0.60$ & $6.20 \pm 1.22$ \\
LBANP (8) & $1.05 \pm 0.11$ & $1.90 \pm 0.13$ & $2.11 \pm 0.09$ & $9.99 \pm 0.42$ & $14.05 \pm 0.06$ \\
LBANP (128) & $0.69 \pm 0.15$ & $1.70 \pm 0.15$ & $2.42 \pm 0.13$ & $10.45 \pm 0.43$ & $14.69 \pm 0.17$ \\
ISANP (8) & $1.11 \pm 0.28$ & $2.01 \pm 0.18$ & $2.60 \pm 0.16$ & $12.95 \pm 0.74$ & $18.18 \pm 0.61$ \\
ISANP (128) & $1.84 \pm 0.37$ & $4.09 \pm 0.61$ & $4.04 \pm 1.58$ & $35.65 \pm 3.04$ & $37.41 \pm 2.01$ \\
ISANP-2 (8) & $0.99 \pm 0.09$ & $3.00 \pm 0.14$ & $6.08 \pm 0.09$ & $23.98 \pm 0.42$ & $34.31 \pm 0.01$ \\
ISANP-2 (128) & $1.01 \pm 0.25$ & $2.38 \pm 0.25$ & $4.71 \pm 0.39$ & $16.14 \pm 0.95$ & $22.77 \pm 0.78$ \\
\bottomrule
\end{tabular}
\end{table}

\textbf{Result:} Table \ref{tab:contextual_bandit_experiment} shows the cumulative regrets of ISANPs and baselines. The replicated baselines have similar performance ranking,  but the results are inconsistent with \citet{feng2022latent} in two ways: 1) the cumulative regrets are generally higher, and 2) increasing the number of latent vectors does not guarantee to enhance the performance of LBANPs. We repeated experiments multiple times using different training datasets and got different results. Thus, we hypothesize the inconsistent performance is caused by NPs' sensitivity to the training dataset for this specific task. For ISANP and ISANP-2, they are not performing consistently across different $\delta$s. When $\delta = 0.7$, ISANP and ISANP-2 are performing comparatively with the baselines, where ISANP-2 even outperforms LBANP when they both use 8 latent vectors. However, ISANPs' performance dropped significantly as $\delta$ increases. The gap is most obvious when $\delta \leq 0.99$.

\subsubsection{Further Elaboration of the Result}
\label{app:contextual_app}
According to the experiment setup, Contextual Bandits' training process encourages models to select the actions associated with high-reward regions. For example, by mistakenly predicting that a point is in a high-reward region, the expected \textit{regret} $= 1.2 - 1.0 = 0.2$. However, by mistakenly predicting that a point is in the low-reward region, the expected \textit{regret} $= 50 - 1.2 = 48.8$, which is much larger. In the evaluation process, $\delta$ will be selected from $[0.7, 0.9, 0.95, 0.99, 0.995]$, which are all larger than the expected $\delta$ during the training process, $0.5$. Therefore, sample points are less likely to fall into the high-reward regions, so if a model keeps selecting high-reward actions, they will gain large cumulative regrets. This is how cumulative regret represents models' abilities to adjust their strategies based on the context data points.

We first ran the experiment with the default configuration and randomly generated training dataset, but the results were way worse than what's shown in the original paper. We contacted \citet{feng2022latent}, and they confirmed that NPs perform inconsistently at times and are sensitive to the training dataset. After using the training dataset they provided, we got the replication results shown in the report, which became comparative to the original paper, as shown in Figure \ref{fig:contextual_bandit_figures}. By observing the figure, we found that some cumulative curves rise rapidly at the beginning and then become stable very soon, which is especially obvious when $\delta = 0.7$. We consider the rapidly rising cumulative regret as a sign of updating strategies and becoming stable as a sign of converging to a strategy. We can see that the updating only happens at the very beginning of the 2000 evaluation steps, and all the NPs do not utilize the following context data points to update their strategies. From this perspective, we can conclude that NPs are not performing ideally in the contextual bandit task.

\begin{figure}[ht]
  \floatconts
  {fig:visualizing_strategies}
  {\caption{Visualizing NPs' initial and eventual strategies}}
  {%
    \subfigure[$\delta=0.7$]{\label{fig:vis_0.7}%
      \includegraphics[width=0.32\linewidth]{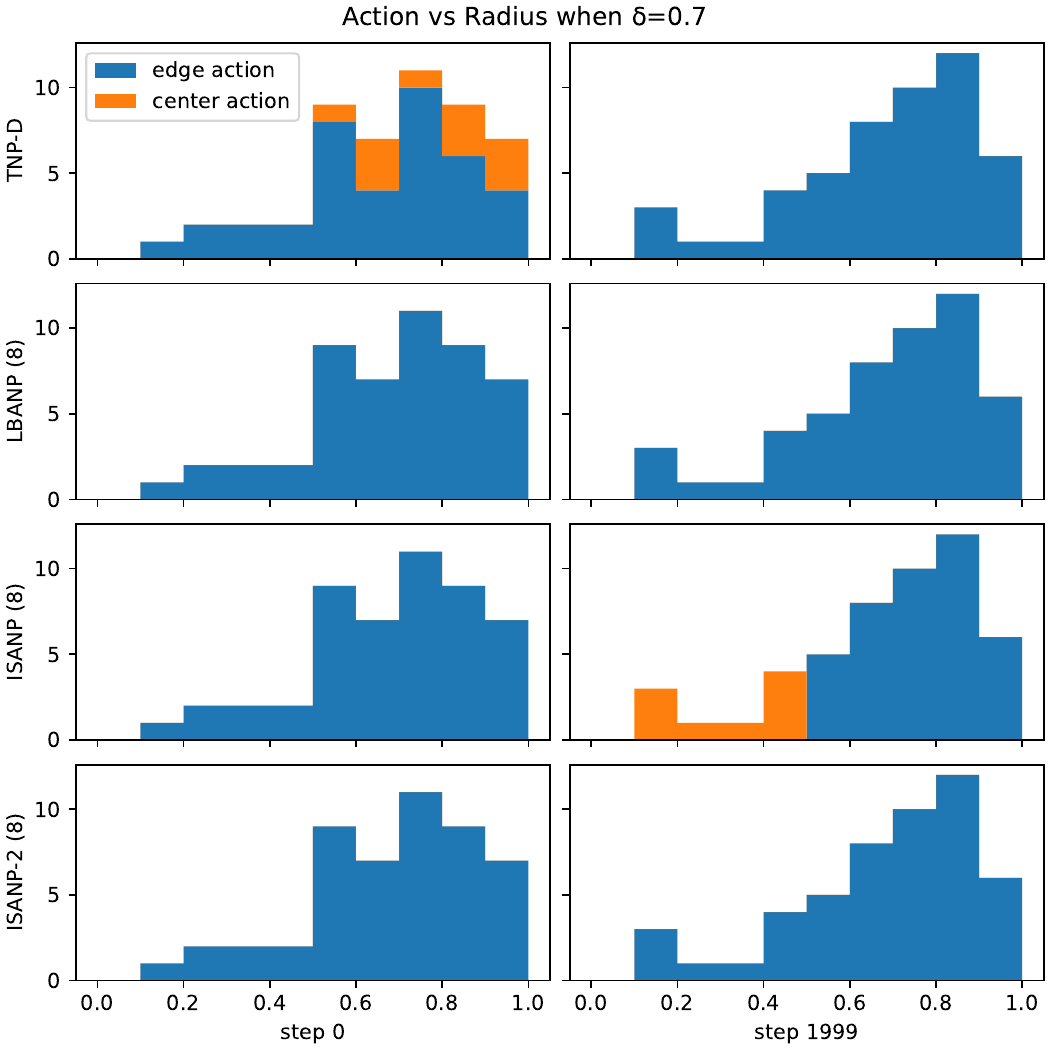}}
    \hfill
    \subfigure[$\delta=0.9$]{\label{fig:vis_0.9}%
      \includegraphics[width=0.32\linewidth]{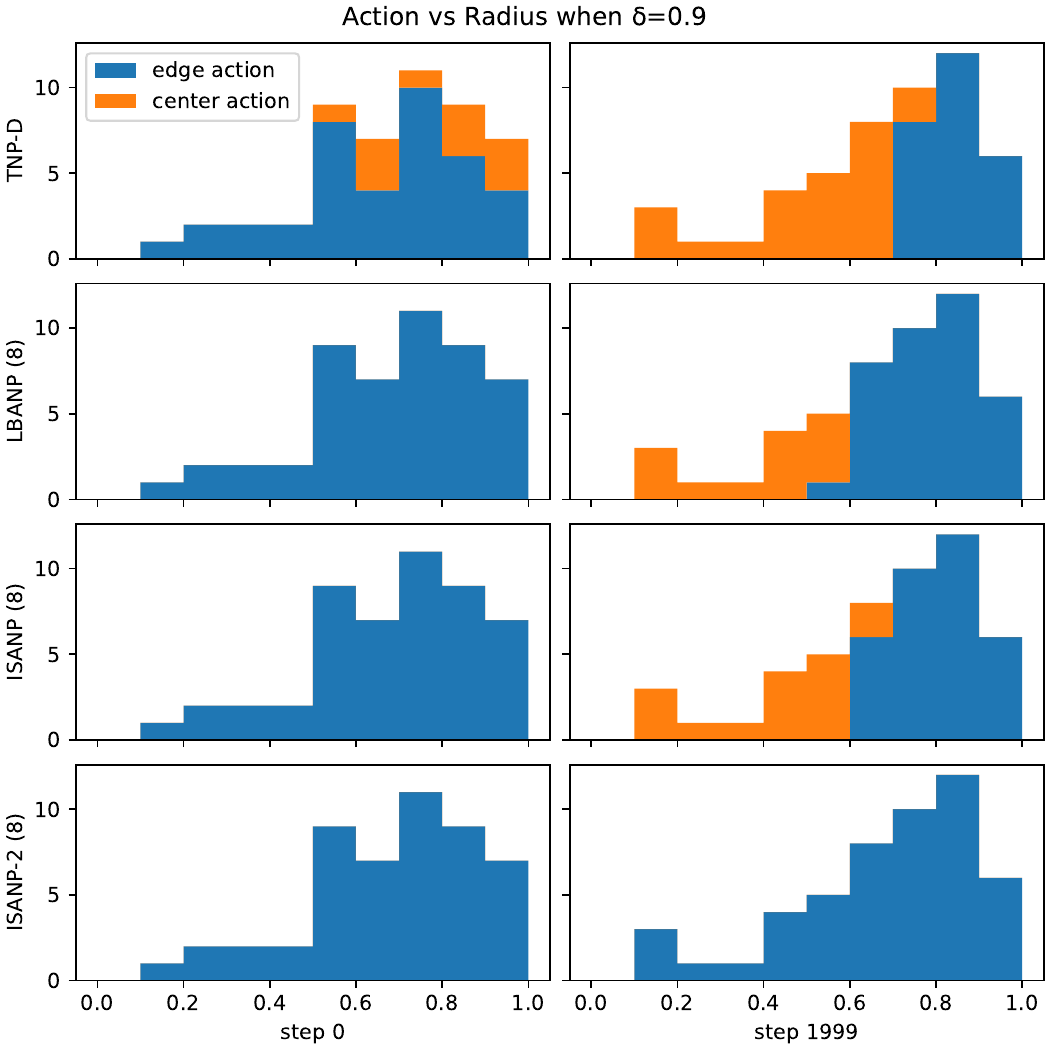}}
    \hfill
    \subfigure[$\delta=0.99$]{\label{fig:vis_0.99}%
      \includegraphics[width=0.32\linewidth]{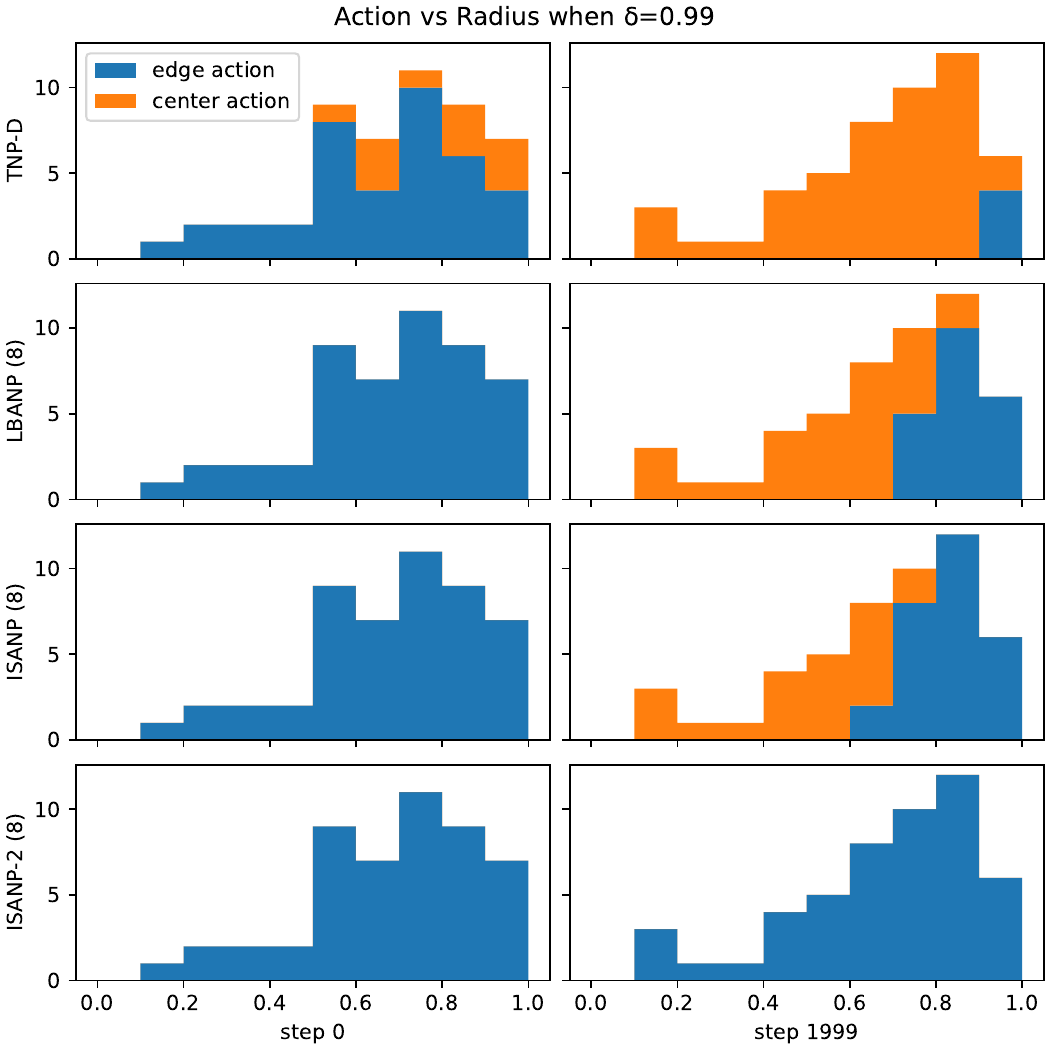}}
  }
\end{figure}

However, the above observation still doesn't explain why the performance of NPs is not consistent. For example, according to Figure \ref{fig:contextual_bandit_figures} and Table \ref{tab:contextual_bandit_experiment}, ISANP and ISANP-2 perform well when $\delta=0.7$ but perform poorly as $\delta$ increases. After some analysis, we hypothesize that cumulative regret might not be a rigorous metric as it doesn't consider NPs' initial strategies at the beginning of the evaluation. Figure \ref{fig:visualizing_strategies} attempts to visualize the strategies used by each NP. The x-axis represents the distance to the center of the disk, and the y-axis represents the frequency; the blue part of a bin represents the points on which NPs predict to take high-reward action, and the orange part represents the points on which NPs predict to take low-reward action. The left column of each subfigure shows four NPs' initial strategies. We can see that NPs tend to select high-reward actions after training, but TNP-D unreasonably prefers to select low-reward actions even when points are far away from the center. The right column of each subfigure shows four NPs' eventual strategies. We can see at $\delta=0.7$, parts of the models, including TNP-D, eventually learned to select high-reward action as much as possible, and it's definitely harder for TNP-D to adjust its strategy to this eventual strategy because it has higher preference to select low-reward actions at the beginning. In contrast, at $\delta=0.99$, the eventual strategies become easier for TNP-D to learn as models should select more low-reward actions here. We think this shows why TNP-D performs worse than other NPs on 'easy' tasks but performs very well on 'difficult' tasks: the 'easy' task is not easy for TNP-D. Therefore, we hypothesize that if the distance between strategies can be measured, \textit{distance(optimal strategy - eventual strategy) - distance(optimal strategy - initial strategy)} is a very important metric to evaluate NPs' ability to learn from context data. Developing such a metric will be a future work. Cumulative regret, however, is not comprehensive enough to represent the changes in strategies.

\begin{figure}[t]
  \floatconts
    {fig:contextual_bandit_figures}
    {\caption{Cumulative regret curves and demo of the uniform disk with different \(\delta\)s.}}
    {
      %--- Row 1 ---
      \subfigure[\(\delta=0.7\) (Regret)]{\label{fig:delta0.7_regret}%
        \includegraphics[width=0.4\linewidth]{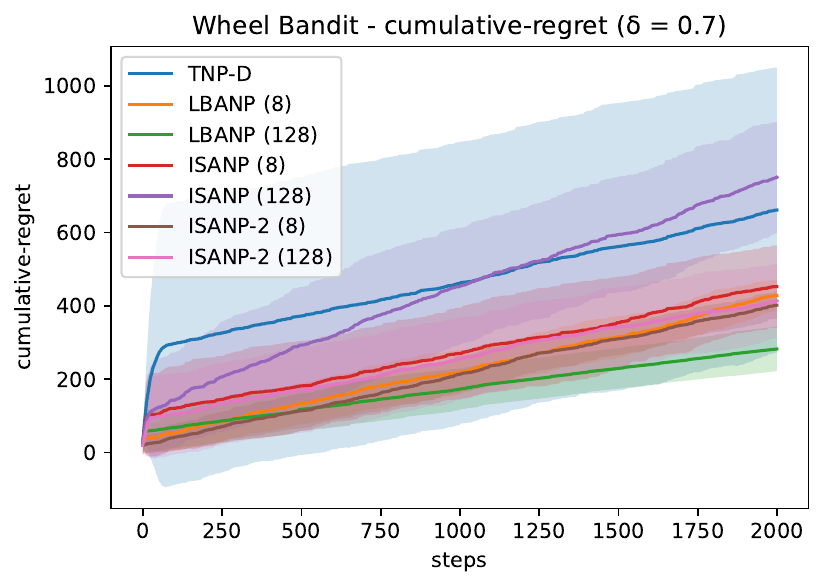}}
      \hfill
      \subfigure[\(\delta=0.7\) (Disk)]{\label{fig:delta0.7_disk}%
        \includegraphics[width=0.4\linewidth]{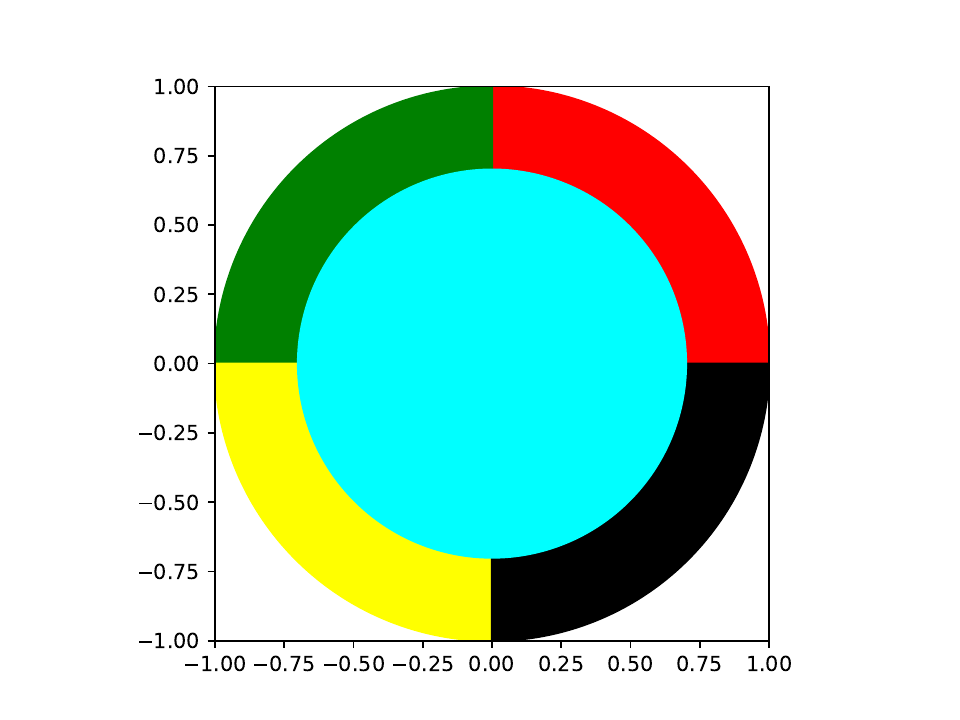}}

      \vfill

      %--- Row 2 ---
      \subfigure[\(\delta=0.9\) (Regret)]{\label{fig:delta0.9_regret}%
        \includegraphics[width=0.4\linewidth]{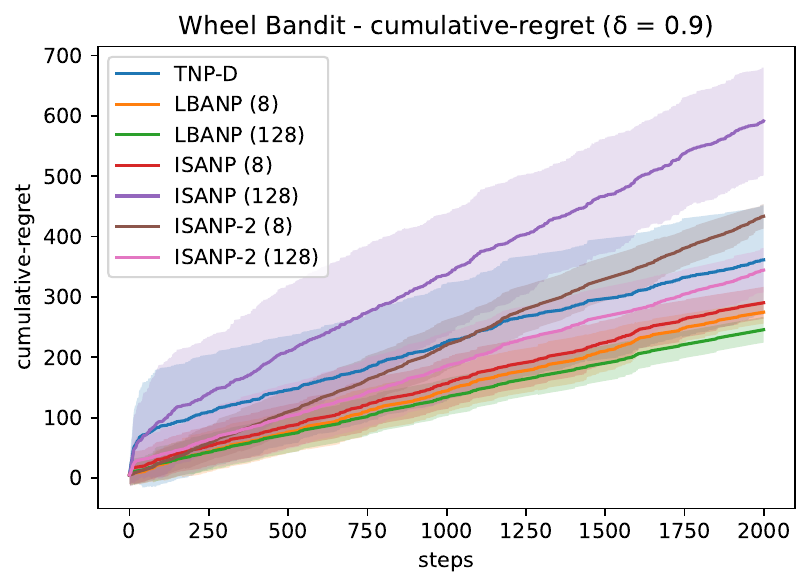}}
      \hfill
      \subfigure[\(\delta=0.9\) (Disk)]{\label{fig:delta0.9_disk}%
        \includegraphics[width=0.4\linewidth]{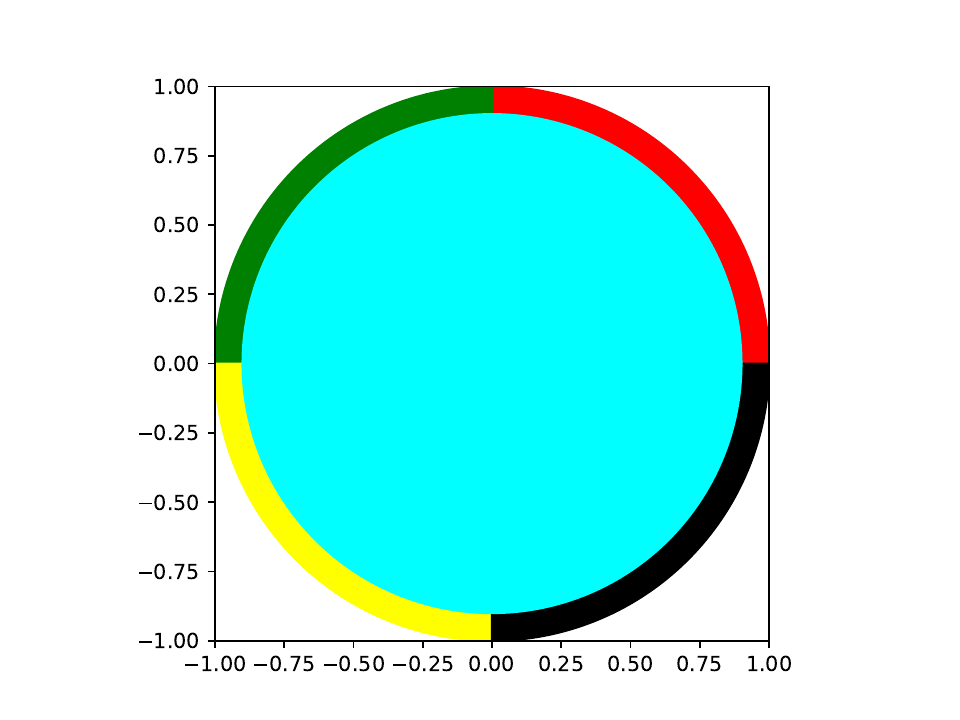}}

      \vfill

      % %--- Row 3 ---
      % \subfigure[\(\delta=0.95\) (Regret)]{\label{fig:delta0.95_regret}%
      %   \includegraphics[width=0.35\linewidth]{images/final_plot_d0.95.pdf}}
      % \hfill
      % \subfigure[\(\delta=0.95\) (Disk)]{\label{fig:delta0.95_disk}%
      %   \includegraphics[width=0.35\linewidth]{images/wheel_delta_0.95.pdf}}

      % \vfill

      %--- Row 4 ---
      \subfigure[\(\delta=0.99\) (Regret)]{\label{fig:delta0.99_regret}%
        \includegraphics[width=0.4\linewidth]{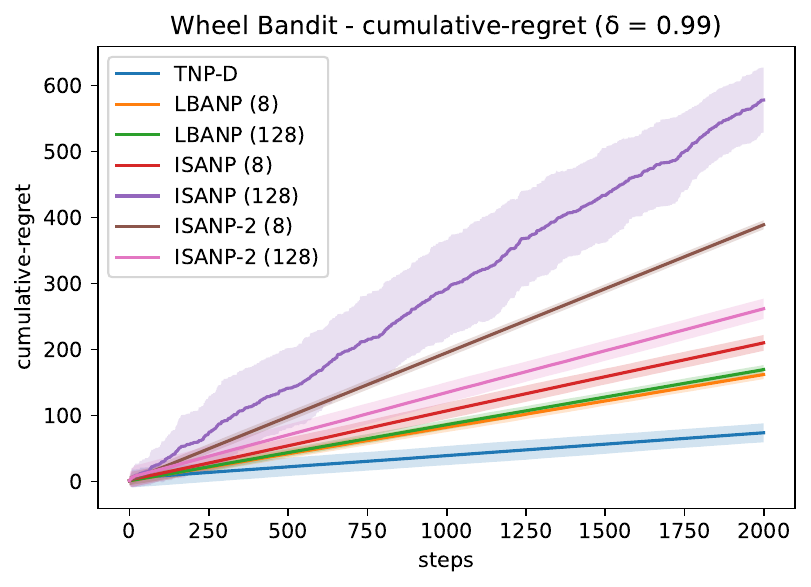}}
      \hfill
      \subfigure[\(\delta=0.99\) (Disk)]{\label{fig:delta0.99_disk}%
        \includegraphics[width=0.4\linewidth]{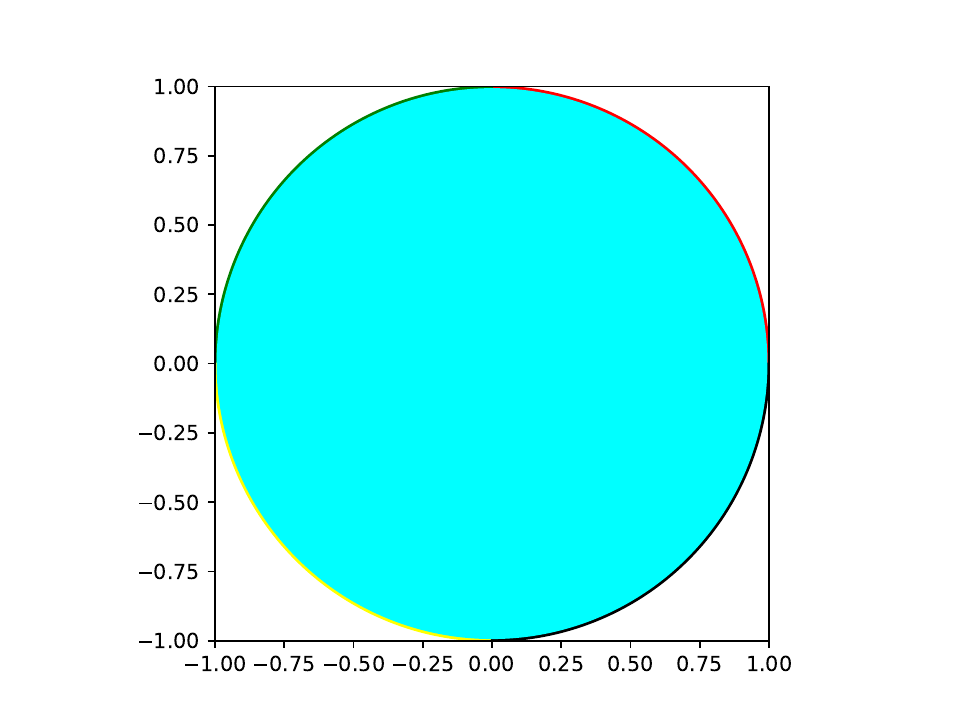}}

      \vfill

      %--- Row 5 ---
      \subfigure[\(\delta=0.995\) (Regret)]{\label{fig:delta0.995_regret}%
        \includegraphics[width=0.4\linewidth]{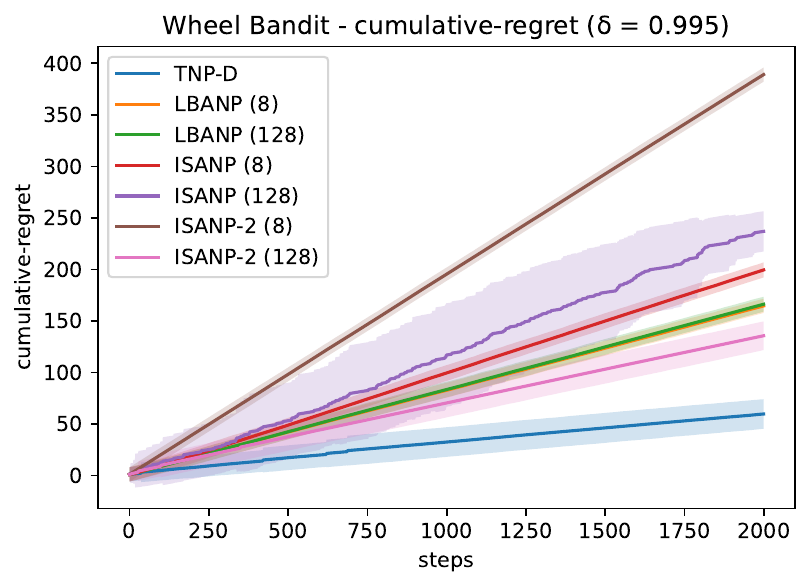}}
      \hfill
      \subfigure[\(\delta=0.995\) (Disk)]{\label{fig:delta0.995_disk}%
        \includegraphics[width=0.4\linewidth]{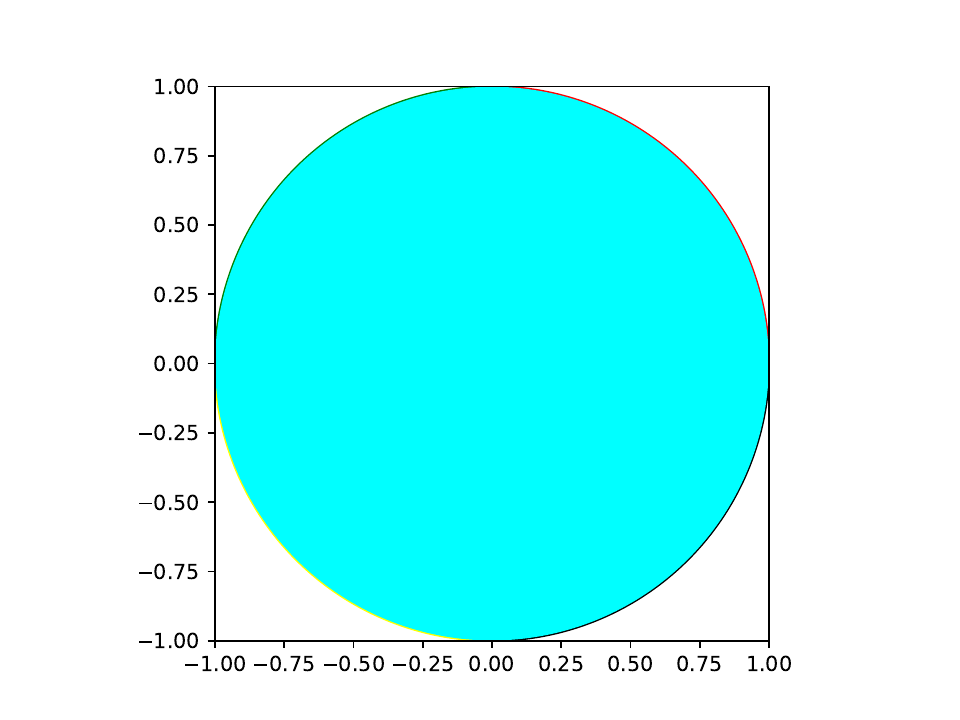}}
    }
\end{figure}

\subsection{Bayesian Optimization}
\label{apd:bayesian_optimization}

\begin{figure}[ht]
    \centering
    \includegraphics[width=\linewidth]{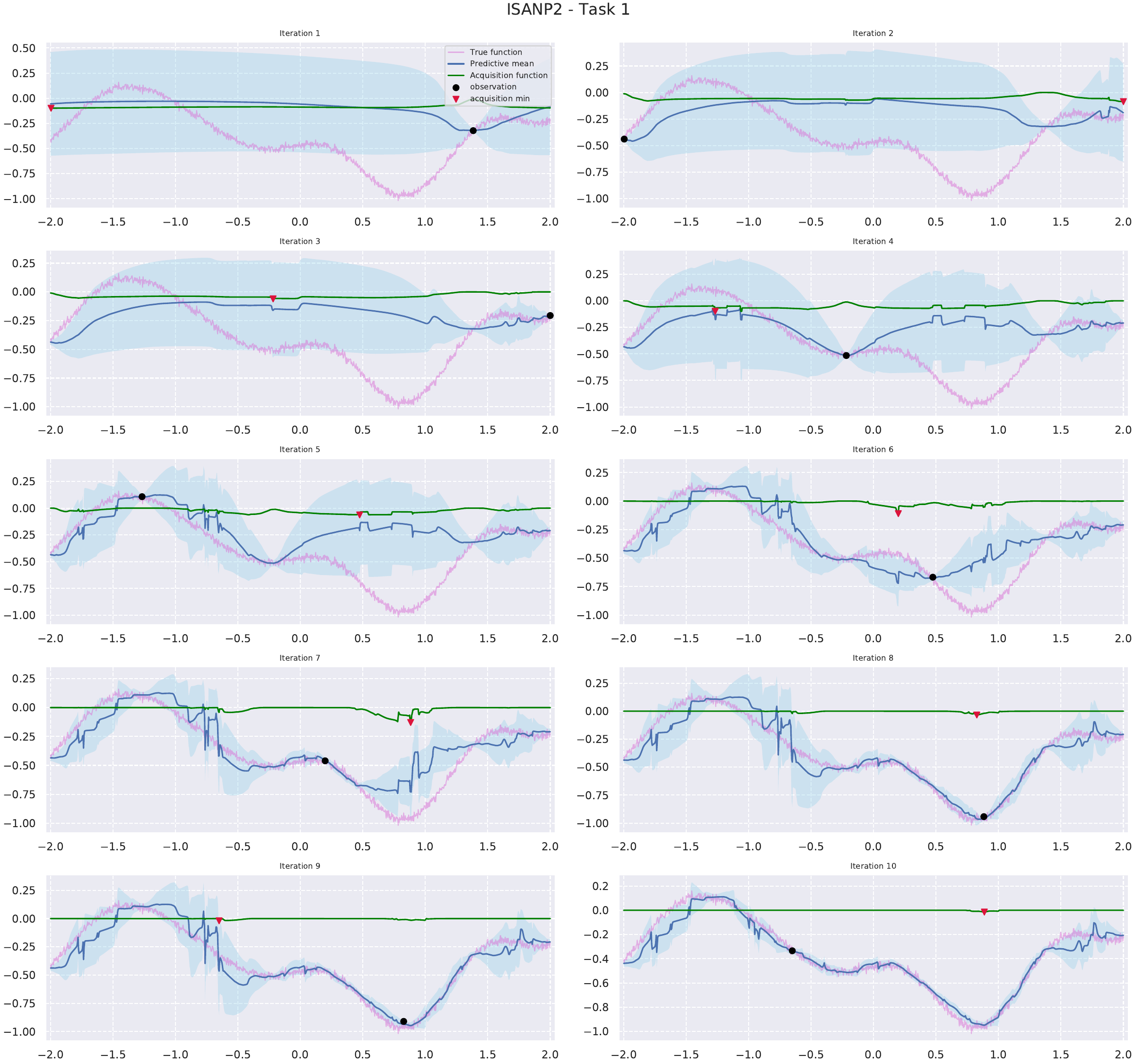}
    \caption{For each iteration, BO uses the new evaluation point and objective function's output to train the surrogate function, specifically ISANP-2 in this demo. Then, it selects the subsequent evaluation point based on the minimum acquisition function value. The predictive mean function approximates the objective function closely in only 10 iterations.}
    \label{fig:bayesian optimization task}
\end{figure}

The goal of Bayesian Optimization (BO) \citep{frazier2018tutorial} is to optimize the evaluation result of a black-box function $f(x)$ without accessing its gradients. The BO procedure iterates continuously, employing a surrogate function to approximate $f(x)$ and utilizing an acquisition function to select the subsequent point for evaluation based on this approximation. A detailed BO demonstration is in Figure \ref{fig:bayesian optimization task}. In this experiment, we use ISANP, ISANP-2, and baselines as surrogate functions and expected improvement criterion (EI) as the acquisition function, testing NPs ability to approximate the 1D objective black-box functions.

The training process of BO is exactly the same as that of 1D regression. During evaluation, 100 objective functions will be respectively generated by GPs with RBF, Mat\'ern 5/2, and Periodic kernels. For each objective function, we run BO for 100 iterations. The means and the standard deviations of the simple regret are reported as evaluation metrics.
\begin{figure}[ht]
    \centering
    \includegraphics[width=\textwidth]{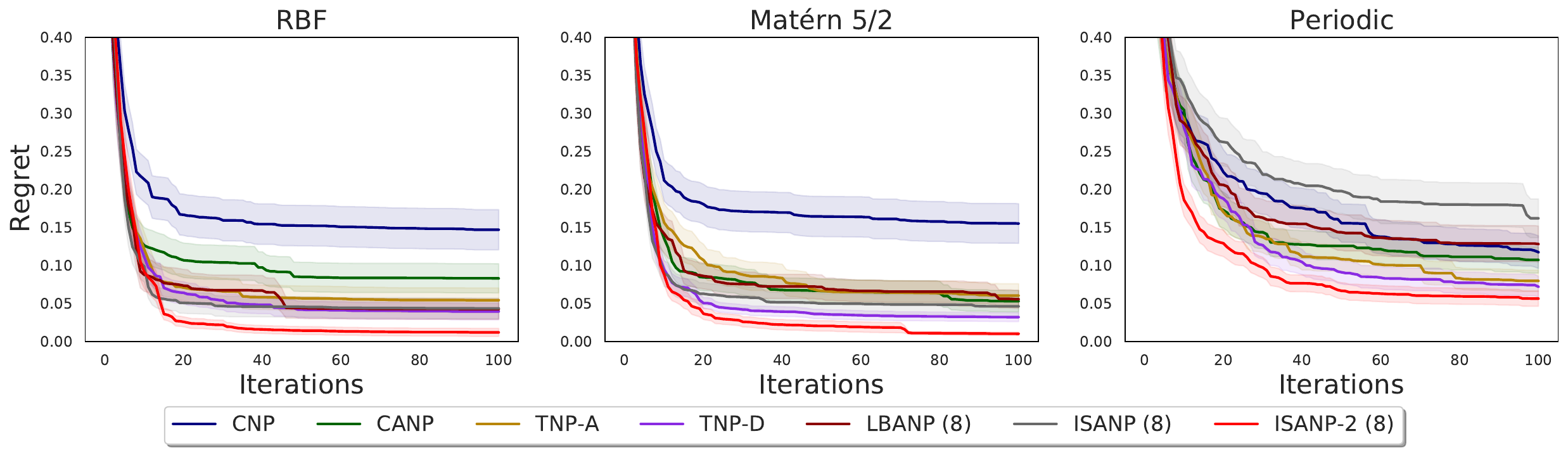}
    \caption{Regret performance on 1D BO tasks (lower is better). For each kernel, we generate 100 functions and report the mean and standard deviation.}
    \label{fig:bayesian optimization}
\end{figure}

\textbf{Result}: As shown in Figure \ref{fig:bayesian optimization}, our replication results have a similar trend to the BO experiment results in \citeauthor{nguyen2022transformer}'s work. Notably, CNP performs way worse than other NPs in approximating objective functions generated by GPs with RBF and Mat\'ern 5/2 kernels. This aligns with our previous discussion about traditional NPs' underfitting issues. For our extension, ISANP-2 with 8 latent vectors outperforms all the other NPs in approximating objective function from all 3 kernels. For ISANP with 8 latent vectors, although performing the worst in estimating periodic objective functions, it performs comparatively with TNP-D in estimating functions generated by GPs with RBF and Mat\'ern 5/2 kernels. BO experiment is not done in \citeauthor{feng2022latent}'s work, and here we can see that LBANP (8) performs comparatively with TNP-D in the RBF kernel task but performs poorly when the objective functions are from unseen kernels. We also planned to run the experiments in a multi-dimensional setting. However, due to the limitation of computing power, we will set it as future work.

\section{Ablation Studies}
\label{apd:ablation_studies}

\begin{figure}[ht]
    \centering
    \subfigure[Log-likelihoods by number of latent vectors on CelebA32.]{%
        \includegraphics[width=0.30\linewidth]{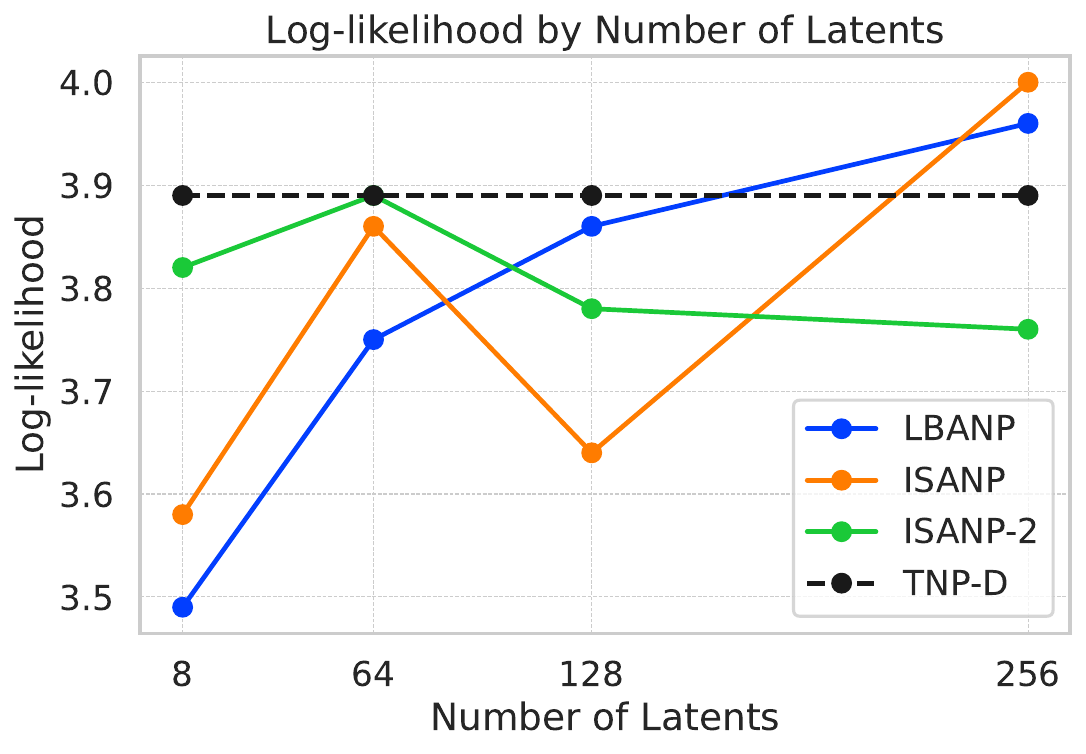}
        \label{fig:latents_performance}
    }
    \hfill
    \subfigure[Empirical Time Complexity]{%
        \includegraphics[width=0.32\linewidth]{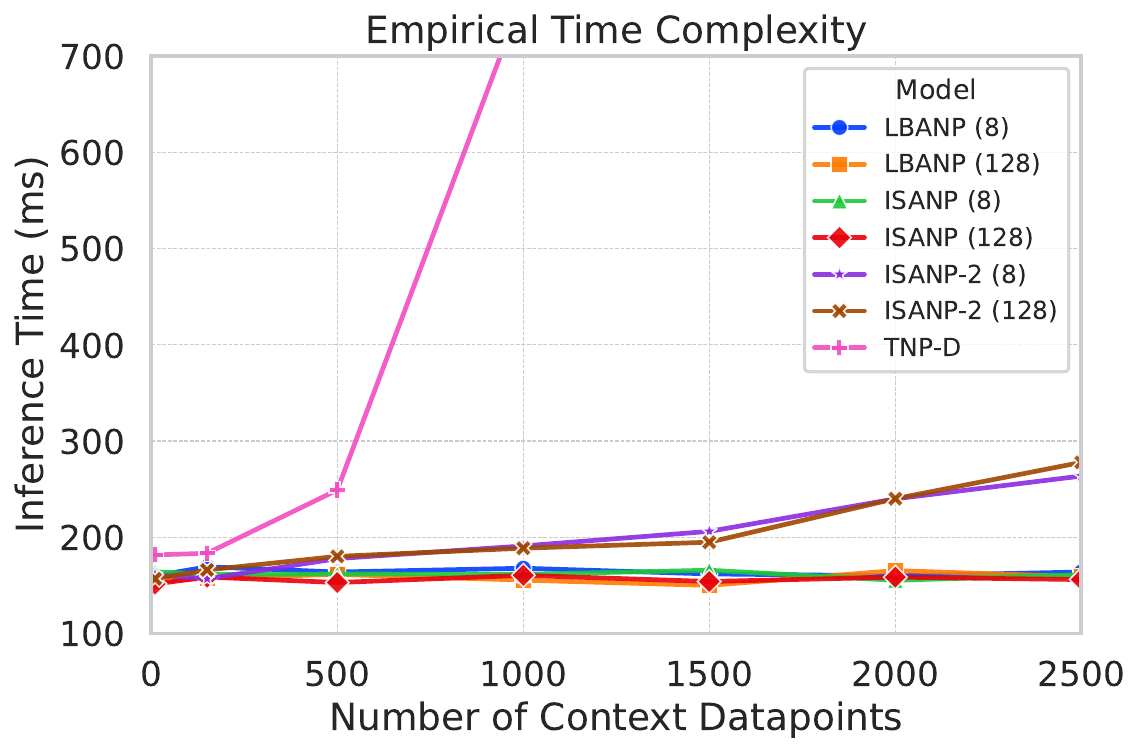}
        \label{fig:time}
    }
    \hfill
    \subfigure[Empirical Memory Complexity]{%
        \includegraphics[width=0.32\linewidth]{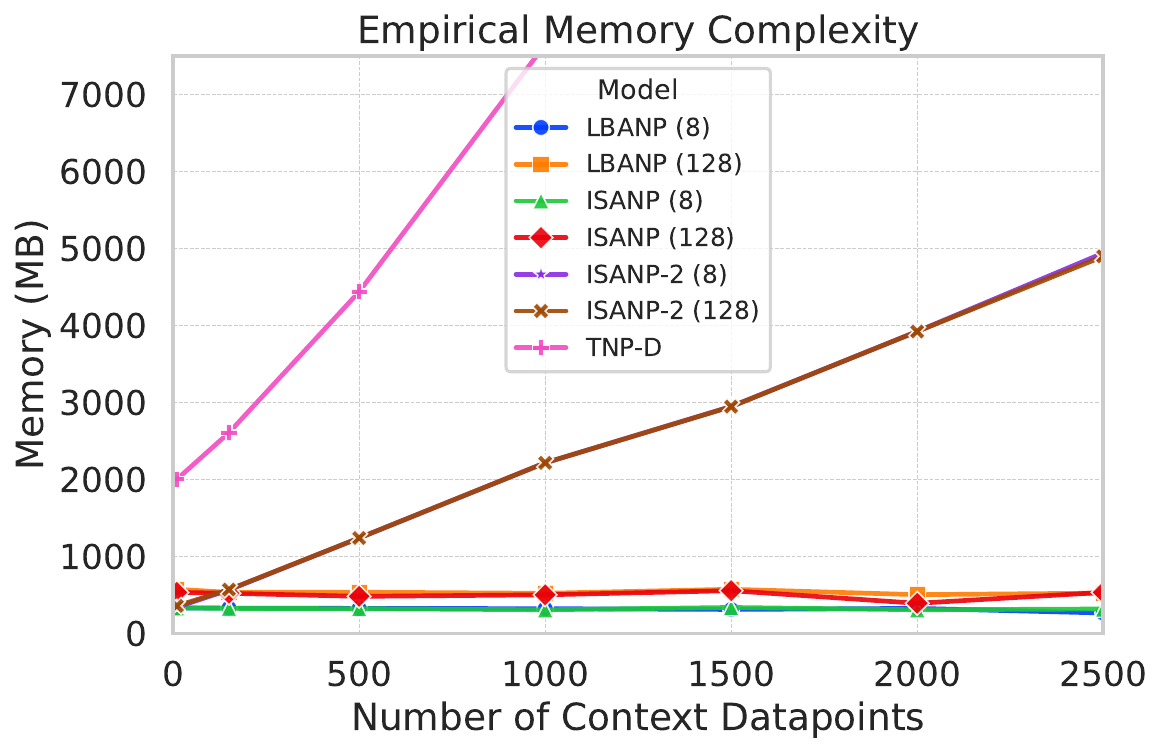}
        \label{fig:memory}
    }
    \caption{Performance and empirical complexity analysis of the models with respect to the number of latent vectors and context datapoints.}
    \label{fig:ablation_studies}
\end{figure}

When using 8 latent vectors ISANP consistently outperforms LBANP in all experiments. We examined the impact of increasing the number of latent vectors from 8 to 256 on the models' performance in the CelebA32 experiment. Figure \ref{fig:latents_performance} reveals that ISANP performance improves, while ISANP-2 does not exhibit any improvement and fails to outperform TNP-D. Indeed, ISANP-2, despite its previous best performance, does not benefit from more latent vectors in processing the context dataset and even shows signs of overfitting. This is likely because ISANP-2 employs direct cross-attention between the target and context datasets, meaning that merely increasing the number of latent vectors does not enrich the context dataset with more useful information for predictions, limiting performance improvement. Interestingly, ISANP shows a performance drop with 128 latent vectors, which is counterintuitive since both LBANP and ISANP are expected to perform better with more latent vectors. We hypothesize that this could be due to variances in the models' log-likelihood, as our analysis is based on a single computation due to resource limitations. Should we conduct multiple experiments on ISANP and LBANP, we anticipate observing a consistent increase in log-likelihood correlating with the number of latent vectors.

%\subsection{Empirical Computational Complexity}
We also investigate how models adapt to larger context sets by varying its size and measuring the empirical time and memory complexity during the prediction of a target dataset with fixed size, using pre-computed context embeddings for the 1D Regression task. Figures \ref{fig:time} and \ref{fig:memory} reveal that TNP-D complexity increases quadratically with the context dataset size, while linearly in the case of ISANP-2. ISANP and LBANP maintain a constant empirical complexity, in line with the theoretical complexity in Table \ref{tab: computational complexity}. This demonstrates the effectiveness of LBANP to maintain constant time and memory complexity during queries while achieving competitive results comparable to TNPs, showing significant improvement and potential for real-world scenarios.

\end{document}